\PassOptionsToPackage{table}{xcolor}

\documentclass[acmsmall,nonacm]{acmart}

\AtBeginDocument{%
  \providecommand\BibTeX{{%
    \normalfont B\kern-0.5em{\scshape i\kern-0.25em b}\kern-0.8em\TeX}}}

\setcopyright{none}              
\renewcommand\footnotetextcopyrightpermission[1]{} 
\acmBooktitle{Woodstock '18: ACM Symposium on Neural Gaze Detection,
 June 03--05, 2018, Woodstock, NY} 
\acmPrice{15.00}
\acmISBN{978-1-4503-XXXX-X/18/06}

\usepackage{tabularx}
\usepackage{makecell}
\usepackage{float}
\usepackage{tabulary}
\usepackage{xcolor} 
\usepackage{comment}
\usepackage{subcaption}
\usepackage[english]{babel}
\usepackage[autostyle]{csquotes}
\usepackage{graphicx} 
\usepackage{amsmath}
\usepackage{multirow}
\usepackage{graphicx}
\usepackage{comment}
\usepackage{makecell}
\usepackage{array}
\usepackage{soul}
\usepackage{lscape}
\usepackage{booktabs}
\usepackage{xcolor}
\usepackage{colortbl}
\usepackage{xfp}

\usepackage[bottom]{footmisc}

\newcommand{\heatmapcolor}[1]{%
    \ifnum \fpeval{#1<3}=1 \cellcolor{yellow!30}%
    \else \ifnum \fpeval{#1<5}=1 \cellcolor{yellow!60}%
    \else \ifnum \fpeval{#1<7}=1 \cellcolor{orange!70}%
    \else \cellcolor{red!90}%
    \fi\fi\fi
    #1%
}
\newcolumntype{?}{!{\vrule width 1pt}}

\newcommand{\quotes}[1]{``#1''}

\begin{document}

\title[How RL can Deliver Desired User Experience in VR Personalized Arachnophobia Treatment.]{Spiders Based on Anxiety: How Reinforcement Learning Can Deliver Desired User Experience in Virtual Reality Personalized Arachnophobia Treatment}

%


\author{Athar Mahmoudi-Nejad}
\affiliation{%
  \institution{University of Alberta}
  \city{Edmonton}
  \state{Alberta}
  \country{Canada}}
\email{athar1@ualberta.ca}

\author{Matthew Guzdial}
\affiliation{%
  \institution{University of Alberta}
  \city{Edmonton}
  \state{Alberta}
  \country{Canada}}
\email{guzdial@ualberta.ca}

\author{Pierre Boulanger}
\affiliation{%
 \institution{University of Alberta}
 \city{Edmonton}
 \state{Alberta}
 \country{Canada}}
 \email{pierreb@ualberta.ca}

\renewcommand{\shortauthors}{Mahmoudi-Nejad, et al.}
\newcommand{\framework}{EDPCGRL}

\begin{abstract}
The need to generate a spider to provoke a desired anxiety response arises in the context of personalized virtual reality exposure therapy (VRET), a treatment approach for arachnophobia. 
This treatment involves patients observing virtual spiders in order to become desensitized and decrease their phobia, which requires that the spiders elicit specific anxiety responses.
However, VRET approaches tend to require therapists to hand-select the appropriate spider for each patient, which is a time-consuming process and takes significant technical knowledge and patient insight. 
While automated methods exist, they tend to employ rules-based approaches with minimal ability to adapt to specific users. 
To address these challenges, we present a framework for VRET utilizing procedural content generation (PCG) and reinforcement learning (RL), which automatically adapts a spider to elicit a desired anxiety response. We demonstrate the superior performance of this system compared to a more common rules-based VRET method.

\end{abstract}



\maketitle

\section{Introduction}


Anxiety disorders are the most prevalent mental illness, with an estimated 301 million people living with them in 2019~\cite{yang2021global}.
Anxiety disorders encompass a range of conditions including panic disorder, agoraphobia, specific phobias, social anxiety disorder, obsessive-compulsive disorder, post-traumatic stress disorder, and generalized anxiety disorder~\cite{yang2021global}.
Anxiety disorders result in notable challenges including decreased well-being, low productivity, reduced quality of life, and frequent medical visits~\cite{ost2023cognitive}.
Cognitive-behavioral therapy is a highly effective and long-lasting psychological treatment for anxiety disorders. It focuses on identifying, understanding, and modifying thinking and behavior patterns~\cite{dobson2021handbook}.
Exposure therapy (ET) is a type of cognitive-behavioral therapy for reducing anxiety and fear responses. In this type of therapy, a person is gradually exposed to a feared situation or object, which decreases the user's sensitivity over time. Researchers have found this type of therapy to be particularly effective for treating obsessive-compulsive disorders and phobias~\cite{kaczkurkin2022cognitive}.
In this work, we focus specifically on technologies for, and the user experience of, a treatment for phobic individuals.

A relatively new form of exposure therapy is Virtual Reality Exposure Therapy (VRET) which has gained popularity for treating anxiety disorders~\cite{carl2019virtual, maples2017use, valmaggia2016virtual}. Compared to traditional exposure therapy, VRET is able to immerse individuals in virtual situations or locations that may not be practical or safe to encounter in real life. For instance, patients with arachnophobia are exposed to virtual spiders to learn to deal with the elicited fear. There is substantial evidence demonstrating a positive outcome from VRET for the treatment of most phobias~\cite{freitas2021virtual}. Phobia-related VRET induces several physiological responses that correspond to anxiety levels in users such as blood pressure, muscle vibration, heart rate, and sweating~\cite{bun2017low, vsalkevicius2019anxiety}. VRET approaches then read these physiological responses to approximate anxiety.
However, VRET applications are mainly non-adaptive, wherein all users experience the same content. 
This is despite the fact that personalized treatment has shown greater benefits for individuals{~\cite{smits2019personalized}}. Adaptive VRET attempts to select specialized content to elicit the user's desired anxiety levels in an individual, offering a more personalized experience. 
Current adaptive VRET is mostly either therapist-guided~\cite{kampmann2016exposure, bualan2020towards} 
or rules-based{~\cite{bian2019design, repetto2013virtual}}.

In therapist-guided methods, the therapist manually selects the content for each individual. This process requires learning a new interface for each VRET system for each new phobia, and predicting the likely anxiety response based on their understanding of the user. Rules-based methods rely on predefined rules to determine content selection. These methods show promise in certain cases where changing the content is straightforward, for example, gradually increasing height in the application of VRET for fear of heights~{\cite{bualan2020ether}}. However, in many cases,  content adaptation is not straightforward, as different individuals are frightened by different attributes of the object of fear, such as the spider in the case of arachnophobia {\cite{lindner2019so}}. We provide further evidence for this effect of different individuals fearing different spider attributes more in the case of arachnophobia in Section~{\ref{sec: Spider Personalization Experiment}}.
In cases like this, for therapists to have precise control over an individual's anxiety response would require designers to manually identify many types of user fear responses ahead of time. 
Therefore, an automatic framework that effectively adapts to individuals to achieve desired anxiety levels without extensive therapist intervention or predefined rules is desirable.
Such an automated framework would likely need to include machine learning to adapt to users based on feedback. 
While this might lead to some poor behavior at first due to learning in comparison to a rules-based approach, with sufficient time we would anticipate the potential to outperform such an approach.

In this study, we present an Experience-Driven Procedural Content Generation (EDPCG) via Reinforcement Learning (RL) framework specifically tailored for VRET targeting arachnophobia. Our primary goal is to develop a system that generates content, such as virtual spiders, that can be dynamically adjusted to match the specific anxiety responses of individuals with arachnophobia.
EDPCG covers approaches that dynamically adapt generated content to achieve a desired user experience~\cite{yannakakis2011experience}. Although EDPCG was originally designed for games, it can be applied to other domains of human-computer interaction (HCI) that involve automated customized content generation~\cite{yannakakis2011experience}.
The generalized EDPCG framework assumes some content generator $G$, which is dependent on a model of the current player $m_{curr}$. This model of the current player is constructed and iteratively refined by observing the player's actions in-game. 
The goal of such a system is to have $G$ produce content based on $m_{curr}$ that satisfies some experiential requirements $E$, such as enjoyment and difficulty.
Example EDPCG components in the context of games include: modeling player experience based on gameplay content and player characteristics, evaluating content quality in relation to the modeled experience, optimizing content representation for better generator performance, and using a content generator to search for ideal content that improves the player's experience according to the developed model {~\cite{yannakakis2011experience}}.
While prior studies~\cite{dimovska2010towards,correa2014new,hocine2015adaptation,i2018toward,lyu2023procedural}
have employed EDPCG in various therapy domains, their methodologies are based on rules-based approaches, limiting the system's generalizability to users with diverse behaviors.

We utilize RL as the content generator $G$ in an EDPCG setup~\cite{guzdial2022reinforcement, khalifa2020pcgrl}, due to its demonstrated effectiveness compared to other EDPCG approaches~\cite{mahmoudi2021arachnophobia}. This EDPCGRL framework can automatically adapt to a user given a desired anxiety level defined by a therapist, which would be our experiential requirement, based on sensors that define our player model.
To the best of our knowledge, our EDPCGRL approach is the first PCGRL application to any real-world rehabilitation task. 
Previously, EDPCGRL has been applied to a simplified synthetic environment, where users were simulated using a mixture of Gaussians. In real-world settings, user behavior is complex and influenced by various factors, such as emotions, past experiences, and individual responses to therapy. This complexity makes adapting to real users a challenging task, requiring models to dynamically adjust to unpredictable behavior patterns. In addition, applying EDPCGRL to real-world settings requires designing an effective real-time interaction framework between users and the environment to approximate an exposure therapy session. 

To address these challenges, we extend EDPCGRL to evaluate its applicability in real-world settings. Specifically, we designed an end-to-end interaction framework to systematically collect and analyze both subjective and objective internal states of users in real-time, enabling an RL agent to observe this information and adapt the environment accordingly.  This interaction framework allows for benchmarking EDPCGRL against state-of-the-art rules-based methods in real-world settings, in terms of their adaptation performance to both subjective and objective internal states of users. 
Our analysis reveals that users respond differently to environmental changes; 
therefore, learning each user's behavior and adapting to it is crucial for personalized treatment.

We focus on arachnophobia as an initial implementation of our EDPCGRL framework since spiders can be easily represented as virtual objects compared to other complex scenarios such as social anxiety. 
While our primary focus is on arachnophobia in this research, we have found commonalities in exposure therapy treatment approaches for different anxiety disorders {\cite{brown2023anxiety}}. This indicates that although our current framework is specifically designed for arachnophobia, there is potential for it to be applied to other anxiety disorders in future research.

In the arachnophobia implementation, the RL agent dynamically generates spiders to induce the desired anxiety level, with the framework aiming to elicit a desired anxiety level for a set period of time.
In practice, therapists would define this desired anxiety level. 
However, in contrast to therapist-guided VRET, they would not need to manually choose the specific spiders to show a user.
Instead, they could simply identify their goals for a user, and the system would automatically adjust to achieve them.

To evaluate the effectiveness of our framework, we conducted two human subject studies. 
Our first study confirmed that humans had a measurable physiological response to our virtual spiders, which was the minimum requirement of our framework.
The second study evaluated our framework in comparison to a rules-based method. 
We found that our framework was effective at achieving desired anxiety levels, indicating that it may be feasible in clinical settings.
However, we only tested the system's ability to reach desired anxiety levels, not its effect on therapeutic outcomes.
Finally, we analyzed our results (see Section~{\ref{sec: Spider Personalization Experiment})} to investigate the benefits of personalizing spiders to each user.

We propose the following contributions:
\begin{itemize}
   \item 
   \textit{Contribution 1 (C1):} 
    Developing an end-to-end personalized EDPCGRL framework, which incorporates real-time human interaction to automatically adapt a VR environment to a user.
   \item \textit{Contribution 2 (C2):} Conducting a human subject study demonstrating that our virtual spiders can elicit anxiety reactions in individuals and impact their physiological measures.
   \item \textit{Contribution 3 (C3):} Conducting a human subject study to evaluate the effectiveness of our framework compared to a rules-based method by collecting both subjective and objective measures during the experiments.
   \item \textit{Contribution 4 (C4):} Demonstrating that different participants exhibit anxiety responses to different attributes of spiders, emphasizing the necessity for personalized treatment for each participant.  

\end{itemize}

The paper is structured as follows: Section {\ref{sec:Related Work}} provides background on physiological measures for estimating anxiety levels, PCG for rehabilitation, VRET for arachnophobia, and adaptive VR using physiological measures.
Section {\ref{sec:Proposed Framework}} presents our proposed framework and Section {\ref{sec:Arachnophobia Implementation}} details the implementation of our framework for arachnophobia as a case study, addressing \textit{C1}. Section {\ref{sec:Requirements Verification Study}} describes our requirement verification study, which includes our first human subject study and covers \textit{C2}. Section {\ref{sec:Evaluating EDPCGRL Framework Study}} focuses on evaluating the EDPCGRL framework, detailing our human subject study that compares our framework to a rules-based method and covers \textit{C3}. Section {\ref{sec: Spider Personalization Experiment}} explores the spider personalization experiment, supporting the importance of personalization in VRET and covering \textit{C4}.
Finally, Section {\ref{sec:Limitations and Future Work}} discusses the limitations of our study and outlines future directions, while Section {\ref{sec:Conclusion}} concludes the paper.
\section{Related Work}
\label{sec:Related Work}

\subsection{Physiological Measures}
\label{RelatedWork: Physiological}
In our research, we need to identify appropriate physiological measures that can be used to assess anxiety levels in real-time.
Past research in the field of anxiety has utilized physiological assessments of autonomic nervous system activity, primarily heart rate variability (HRV)~\cite{cheng2022heart, harrewijn2018heart, alvares2013reduced, chalmers2014anxiety, gaebler2013heart}, and electrodermal activity (EDA)~\cite{christian2023electrodermal, kimani2019addressing, kritikos2019anxiety, lonsdorf2017don, vsalkevicius2019anxiety, sevil2017social, nikolic2018bumping, owens2015can} to evaluate anxiety and emotional responses.
EDA refers to the autonomic changes in the electrical properties of the skin.
EDA recordings consist of two primary components: a phasic component (Skin Conductance Response or SCR) and a tonic component (Skin Conductance Level or SCL). SCRs, characterized by rapid fluctuations (typically isolated using high-pass filtering above 0.05 Hz and amplitude thresholds of 0.01-0.04 µS), are directly linked to specific stimuli or events. In contrast, SCL exhibits slower, more gradual changes in response to sustained stimuli or background arousal levels{~\cite{boucsein2012electrodermal}}.


Several studies have demonstrated a strong correlation between SCL and self-reported anxiety in VRET~\cite{wilhelm2005mechanisms, muhlberger2008virtual}, and frequently employed SCL as an anxiety indicator in VRET research~\cite{muhlberger2001repeated, shiban2015fear, wangelin2013enhancing, shiban2013effect, martens2019feels, diemer2016fear}. In this paper, we employed SCL to assess participants' anxiety.

SCR amplitudes have also been applied to approximate anxiety, stress, and arousal. 
Previous studies have extensively utilized SCR amplitudes metrics, including numbers of peaks, mean, or sum of SCR amplitudes, in research focusing on stress ~\cite{visnovcova2016complexity, correia2023effects, bhoja2020psychophysiological, brouwer2018comparison} and anxiety~\cite{suso2019virtual, pick2018autonomic, yee2015insecure, albayrak2023fear, sevincc2018language}.
In this paper,
we utilized SCR to further investigate changes in participants' anxiety.



\subsection{Procedural Content Generation (PCG) for Rehabilitation}
Our proposed framework is built using Procedural Content Generation (PCG) as the underlying structure of an adaptive environment. 
PCG utilizes algorithms to automatically create or modify content. Like in our work, several studies have employed PCG for adaptive rehabilitation purposes. For instance, ~\citet{dimovska2010towards} developed a ski-slalom game to improve balance and persistence by placing gates based on player performance. ~\citet{correa2014new} created a self-adaptive first-person shooter game for amblyopia (lazy eye), adjusting the game parameters using PCG according to the performance of the patient. ~\citet{hocine2015adaptation} proposed a system that adjusted the difficulty of a pointing task for upper limb rehabilitation based on the patient's motor ability and performance. ~\citet{i2018toward} designed a VR labyrinth to enhance emotional self-awareness, procedurally generating the maze and adapting audiovisual elements to induce different emotional states. ~\citet{lyu2023procedural} utilized PCG to assist physiotherapists in customizing gait rehabilitation based on patient precision and efficiency, paralleling our goal for a system to help therapists treating phobias.

These studies employed constructive or rules-based PCG methods~\cite{guzdial2022classical} to generate diverse graphical content, relying on predefined rules under the assumption of known or uniform subject responses. In contrast, our approach assumes unknown subject responses, where we do not assume all subjects will respond to all spiders identically, requiring dynamic content generation and adaptation.
Hence, we propose utilizing PCG via reinforcement learning~{\cite{guzdial2022reinforcement}} to dynamically modify content, avoiding these assumptions. This method allows for adaptive content generation and adjustments based on user needs, promoting dynamic and personalized content creation.

\subsection{Virtual Reality Exposure Therapy (VRET) for Arachnophobia}
In this study, we propose an adaptive framework for Virtual Reality Exposure Therapy (VRET) specifically designed for treating arachnophobia.
VRET interventions for arachnophobia have demonstrated significant reductions in fear of spiders, avoidance behavior, and self-reported fear among clinical populations~\cite{garcia2002virtual, hoffman2003interfaces, bouchard2006effectiveness, cote2005documenting, cote2009cognitive, michaliszyn2010randomized, plasencia2018interactive, minns2019immersive, miloff2019automated, lindner2020gamified, dyrdal2022virtual}.

\citet{garcia2002virtual} was one of the first studies that showed the effectiveness of VRET for arachnophobia, revealing a significant improvement in 83\% of the VRET group compared to 0\% of the waiting list control group.
\citet{michaliszyn2010randomized} and \citet{miloff2019automated} compared VRET with traditional in-vivo exposure therapy and showed significant improvement in both groups.
Most of these studies did not include physiological measures; however, ~\citet{cote2005documenting, cote2009cognitive} collected physiological measures, specifically heart rate variability. They observed significant changes in physiological measures after treatment, indicating a noteworthy reduction in anxiety levels. This study provides evidence that physiological measures can serve as a means of quantifying anxiety in VRET.

Most of these studies~\cite{garcia2002virtual, hoffman2003interfaces, bouchard2006effectiveness, cote2005documenting, cote2009cognitive, michaliszyn2010randomized, minns2019immersive, miloff2019automated, lindner2020gamified, dyrdal2022virtual} pre-design various scenarios. These scenarios involve attributes related to spiders, including size, behavior, distance, and interaction patterns (e.g., observation or tactile contact). Participants progress through these scenarios incrementally, advancing to the next scenario upon completing the current one.
\citet{plasencia2018interactive} designed an augmented reality exposure therapy for arachnophobia, involving four treatment stages: static spider, one moving spider, multiple moving spiders, and many moving spiders. Assignment to these stages was based on pre-treatment assessments using the Fear of Spider Questionnaires (FSQ).
They employed a video game-like setting for their study procedure, similar to our studies. 
However, they employed the same framework for all participants without accounting for individual differences, thus assuming uniform responses.


~\citet{kritikos2021personalized}, to the best of our knowledge, is the only prior work including an adaptive VR environment for individuals with arachnophobia. They employed a set of rules to control the appearance and behavior of spiders, aiming to induce the desired level of anxiety in participants. The level of anxiety experienced by participants was measured using normalized electrodermal activity (EDA) changes, allowing an objective assessment of anxiety levels during VR exposure. 
While their approach was adaptive, it still assumed uniform responses across participants, with the rules increasing or decreasing all spider attributes based on the EDA measures rather than altering individual attributes based on a specific user's response. 
Drawing inspiration from this research, we incorporated EDA as a physiological measure and adapted their rules-based method in our human-subject study.

\subsubsection{Questionnaires and Tests}
Most of the research on VRET for arachnophobia consistently utilizes specific measures, including spider fear, anxiety, and VR-related assessments. Three commonly employed questionnaires for spider fear include the Fear of Spider Questionnaire (FSQ)~\cite{szymanski1995fear}, Spider Phobia Questionnaire (SPQ)~{\cite{klorman1974psychometric}}, and Spider Beliefs Questionnaire (SBQ)~\cite{arntz1993negative}. Additionally, the Behavioral Avoidance Test (BAT)~{\cite{ost1991one}} assesses an individual's avoidance behavior and emotional response when confronted with real spiders. The Subjective Units of Distress Scale (SUDs)~{\cite{wolpe1990practice}} is also used to measure stress/anxiety intensity during exposure, using a self-assessment scale ranging from 0 (no distress or total relaxation) to 100 (highest anxiety/distress ever felt).
For evaluating VR experiences, studies employ the Simulator Sickness Questionnaire (SSQ)~{\cite{kennedy1993simulator}} and Presence Questionnaire (PQ)~\cite{witmer1998measuring, schubert2001experience}.
Following on common trends in this research area, we also drew on FSQ, SUDs, SSQ, and PQ in our study.

\subsection{Adaptive Virtual Reality (VR) using Physiological Measures}

In this study, we propose a framework that adapts the VR environment based on anxiety levels estimated through physiological measures. There has been research on adapting VR environments in real-time using physiological measures to help individuals with different health conditions, which is relevant to our approach.
For instance, ~\citet{fominykh2018conceptual} created a virtual beach environment that alternated between calm and stormy scenarios based on heart rate variability, for adolescents with tension-type headaches. ~\citet{bian2019design} designed an adaptive VR driving simulator for individuals with autism spectrum disorder, which modified task difficulty based on engagement level using participants' physiological responses. ~\citet{dash2019design} designed an adaptive grasping task using audio-visual feedback based on Electromyography (EMG) measurements, specifically tailored for individuals with neurological disorders. ~\citet{tu2018breathcoach} introduced an in-home breathing training system that utilizes physiological feedback through a VR game to improve breathing patterns and user experience. \citet{repetto2013virtual} and \citet{gorini2010virtual} developed a VR tropical island for the treatment of Generalized Anxiety Disorders (GAD), where visual cues such as fire intensity and water movement were adjusted based on the user's heart rate.

While previous studies have shown the effectiveness of adapting VR environments compared to static ones, they often rely on pre-designed rules that may not be applicable to everyone. In our work, we overcome this limitation by dynamically adapting the VR environment for each individual without prior assumptions or knowledge via Reinforcement Learning (RL).
To the best of our knowledge, we are the first to apply reinforcement learning to VRET.

\section{Proposed Framework}
\label{sec:Proposed Framework}

\begin{figure}[tb]
\centering
\includegraphics[width=0.7\columnwidth]{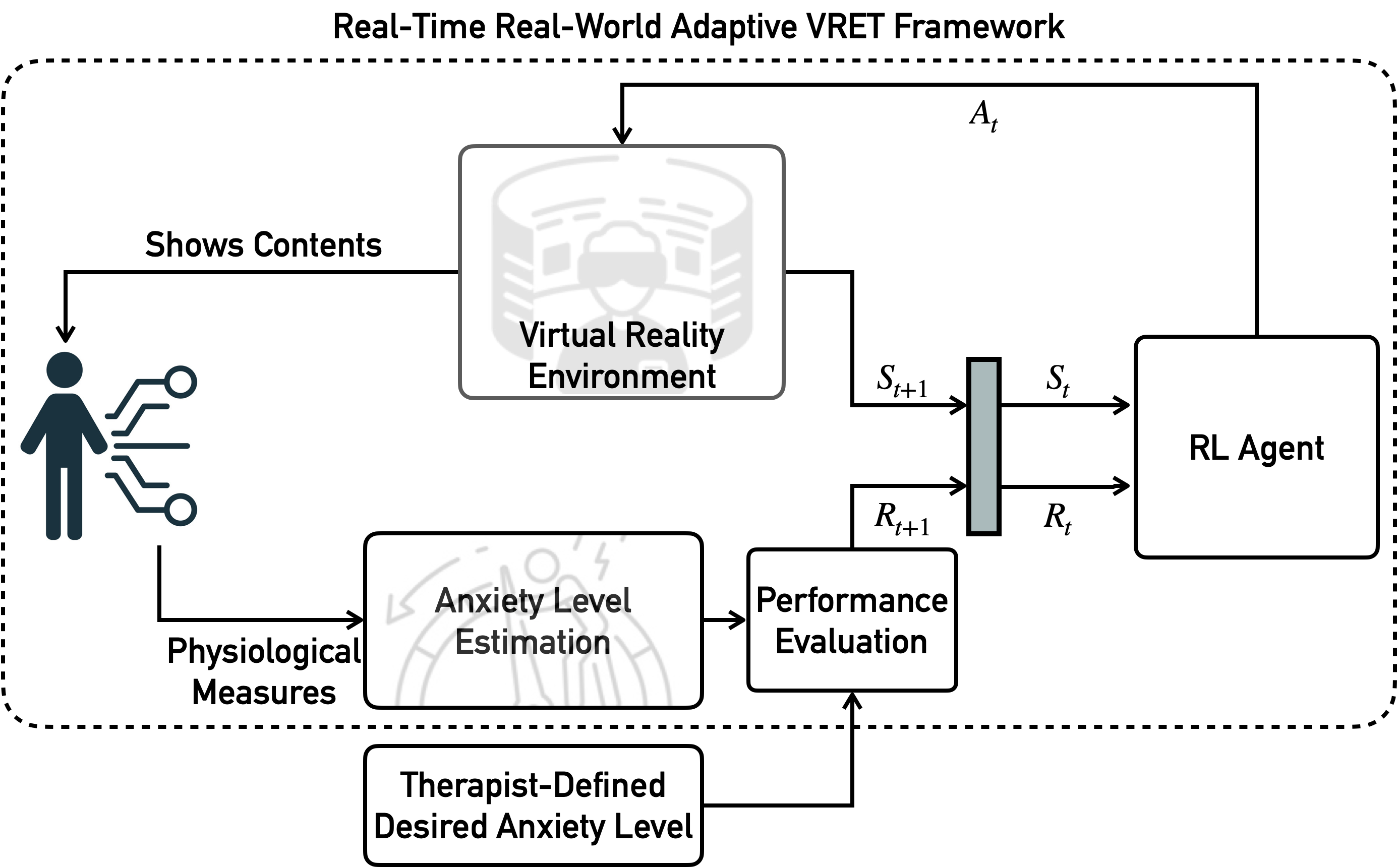}
\caption{Our EDPCGRL framework tailored to real-time, real-world adaptive VRET. In this variation of EDPCGRL, a user interacts with a VR environment while their physiological measures are collected to estimate their anxiety level. Based on the estimated and desired anxiety levels comparison via the Performance Evaluation component, an RL agent determines how to modify the VR environment.}
\label{fig: system}
\Description{
The Figure presents a circular flow diagram including a human figure and three primary components: the Virtual Reality Environment, Stress Level Estimation, and PCGRL Agent. The flow is represented by a series of arrows connecting these elements as follows:
\begin{itemize}
    \item From the Virtual Reality Environment to the human, indicating immersion or interaction.
    \item From the human to the Stress Level Estimation, signifying the collection of physiological measures.
    \item From the PCGRL Agent to the Virtual Reality Environment, denoting the next action.
    \item From the Stress Level Estimation to the PCGRL Agent, representing the reward.
    \item From the Virtual Reality Environment to the PCGRL Agent, including the state information.
\end{itemize}
}
\end{figure}

In this section, we present an EDPCGRL framework tailored to real-time real-world adaptive VRET, denoted as \framework, shown in Figure \ref{fig: system}, for the general VRET case. 
The EDPCGRL framework was initially introduced in~\cite{mahmoudi2021arachnophobia}, where it was evaluated exclusively through simulation. In the current study, we extend this framework to real-time, real-world adaptive VRET. This extension allows us to empirically validate it with real human participants and directly compare its performance against an established state-of-the-art rules-based method.

The box at the top of Figure~\ref{fig: system} represents the Virtual Reality (VR) environment. This environment includes adaptive variables which are adjustable parameters that can alter the user's anxiety level, such as height for someone with a fear of heights. 
In the VRET context, these variables must be relevant to exposure therapy goals. For example, for fear of public speaking disorders, the size of the audience could be an exposure parameter, but the environment’s
brightness is not relevant.

The anxiety estimation acts as a performance measure that quantifies the effect of the user’s interaction with the VR environment. It estimates the user's anxiety level in real-time and compares it to the desired anxiety level to determine a reward for the system. Ideally, a therapist would explicitly define the desired anxiety level for the purpose of exposure therapy. 

The PCGRL component serves as a content generator, automatically adjusting the adaptive variables based on the system reward. Outside the reward, it also takes in the current state, which includes the adaptive variables. The PCGRL component then takes an action that modifies the adaptive variables in order to increase its reward.
The PCGRL component initially takes actions (makes changes) at random, but we have found it quickly learns to take only those actions that will help achieve the desired anxiety level. 
We present an illustrative example in Section {\ref{sec: rule-based}.}

Overall, this framework offers a flexible and adaptable approach to VRET, allowing for personalized therapy for different anxiety disorders. With this foundation, we explore how this framework was specifically tested in the treatment of arachnophobia.

\section{Arachnophobia Implementation}
\label{sec:Arachnophobia Implementation}

For this work, a virtual reality environment was created with a 3D spider model. 
Our adaptive variables were identified based on prior work into attributes of spiders feared by arachnophobic individuals~\cite{lindner2019so}. ~\citet{lindner2019so} asked spider-fearful participants (n=194) to rate seven attributes (grouped into either movement or appearance categories) of spiders influencing their fear response. From these attributes, we selected six for our virtual spider based on their research, shown in Table~\ref{Table: spider_attr}. 
We excluded only the Realness attribute (a binary measure whether the spider was real or not), as our work focused on virtual spiders. 
We adapt the work of~\citet{lindner2019so} for both this and several other aspects of our implementation.

For each attribute, we defined 2-3 ordinal values. 
The movement attributes were Locomotion, Amount of Movement, and Closeness, which respectively describe how the spider moves (specifically the movement of its legs), how much it moves, and its distance to the user. The appearance attributes were Largeness, Hairiness, and Color, indicating the spider's size, whether the spider is hairy or not, and the color of the spider. Note that we specified ordering of the possible values for color, which facilitates the implementation of our adaptation methods. However, this ordering may not accurately reflect real-world scenarios.

\begin{table}
\small
\centering
\caption{The spider attributes \cite{lindner2019so} and defined possible values.}
\label{Table: spider_attr}
\begin{tabular}{ccccccc}
\toprule
\multicolumn{1}{l}{} & \multicolumn{3}{c}{Movement Attributes} & \multicolumn{3}{c}{Appearance Attributes} \\ 
Attribute & Locomotion & \multicolumn{1}{c}{Amount of Movement} & \multicolumn{1}{c}{Closeness} & Largeness & \multicolumn{1}{c}{Hairiness} &
\multicolumn{1}{c}{Color} \\ 
\toprule
\begin{tabular}[c]{@{}c@{}}Possible\\ Values\end{tabular} & \multicolumn{1}{l}{\begin{tabular}[c]{@{}l@{}}0: Standing\\ 1: Walking\\ 2: Jumping\end{tabular}} & \begin{tabular}[c]{@{}l@{}}0: Slightly\\ 1: Moderate\\ 2: Too much\end{tabular} & \begin{tabular}[c]{@{}l@{}}0: Far away\\ 1: In the middle\\ 2: Very close\end{tabular} & \multicolumn{1}{l}{\begin{tabular}[c]{@{}l@{}}0: Small\\ 1: Medium\\ 2: Large\end{tabular}} & \begin{tabular}[c]{@{}l@{}}0: Without\\ 1: With\end{tabular} & \begin{tabular}[c]{@{}l@{}}0: Gray\\ 1: Red\\ 2: Black\end{tabular} \\ 
\bottomrule
\end{tabular}
\end{table}

In this work, we estimated user anxiety by measuring Skin Conductance Level (SCL). We discretized and mapped this value to a scale from 0 to 10, to represent a user's current anxiety level. We compared the current anxiety level to a desired anxiety level to calculate a reward. We structured the reward function as a normal distribution, scaled to the range of $[-1,1]$. The desired anxiety level producing a reward of 1, with the reward reduced exponentially as the anxiety level deviated from the desired one.

The PCGRL Agent component serves as an automatic content generator that adjusts the spider's attributes based on the user's estimated anxiety level. Following the framework of ~\citet{shaker2016procedural} we consider the process of selecting a candidate spider from the space of possible spiders defined by our attributes, as a PCG problem. The state representation is the value of each of the spider's attributes. The PCGRL component then generates an action that increases or decreases one spider's attribute at a time. We employed the tabular Q-learning algorithm with an epsilon-greedy ($\epsilon = 0.05$) action selection policy~\cite{sutton2018reinforcement}. 
We found this to be sufficient given the relatively small number of states (possible states correspond to different spiders with unique attributes. We considered 5 attributes with 3 possible values each, and 1 attribute with 2 possible values, resulting in a total of $3 \times 3 \times 3 \times 3 \times 2 \times 3 = 486$ possible states.).

\section{Requirements Verification Study}
\label{sec:Requirements Verification Study}

In this initial study, we aimed to test whether our VR environment met the fundamental requirements for our framework. These fundamental requirements were (1) whether our virtual spiders could cause human anxiety responses and (2) whether we could measure the changes to their physiological responses. We require (1) as otherwise it is impossible to use a system for virtual reality exposure therapy, and we require (2) as otherwise it is impossible to apply EDPCG. We hypothesized that our VR environment met both requirements. To validate this hypothesis, we conducted a human-subject study with 23 non-arachnophobic individuals.

\begin{figure*}[t]
  \centering
  \begin{subfigure}[b]{0.45\textwidth}
    \includegraphics[width=\textwidth]{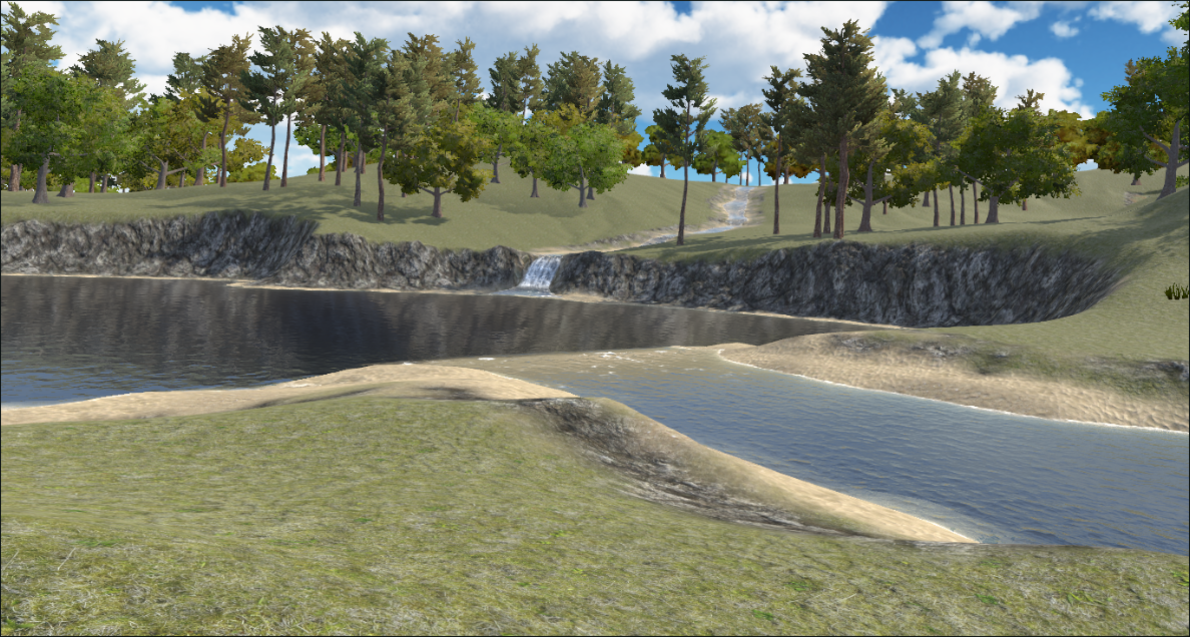}
     \caption{Relaxing} 
     \label{fig: relaxing}
     \Description{Depicts a natural environment with grass and trees, complemented by a river flowing through. The sky above is blue and sunny with a few clouds.}
  \end{subfigure}
  \begin{subfigure}[b]{0.457\textwidth}
    \includegraphics[width=\textwidth]{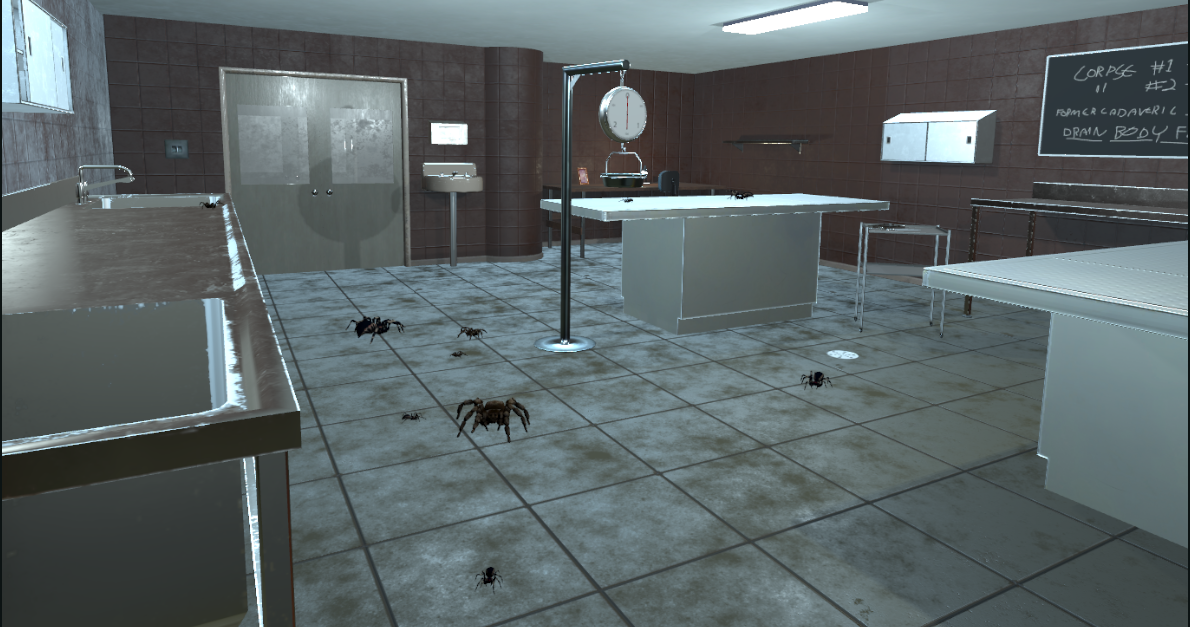}
    \caption{Anxious}
    \label{fig: stressful}
    \Description{The image displays a morgue populated with spiders of varying sizes. Two stainless steel tables are placed in the room. Near the left wall is a countertop equipped with a sink and faucet. On the right wall, there's a blackboard with some text. The ground is covered with tiles that appear dirty and worn. At the very end of the room, directly in the viewer's line of sight is a large door.}
  \end{subfigure} \hfill
  \caption{Developed VR environments for requirements verification study. (a) A relaxing environment featuring natural elements such as mountains, rivers, waterfalls, and trees, accompanied by the sounds of birds. (b) An anxious environment depicting a morgue infested with spiders of varying sizes and colors, accompanied by eerie sounds.}
\end{figure*}

The study used a within-subjects design and presented two environments in a fixed order. Each participant was first immersed in a virtual environment (relaxing environment) designed to induce relaxation for a duration of seven minutes. This environment included natural elements like mountains, rivers, waterfalls, and trees, as well as the sounds of birds (Figure~{\ref{fig: relaxing}}). Afterwards, participants were exposed to another virtual environment (anxious environment) intended to evoke anxiety for seven minutes. This environment depicts a morgue infested with spiders of varying sizes and colors, accompanied by eerie sounds (Figure~{\ref{fig: stressful}})\footnote{Note that the relaxation-first order follows existing practices{~\cite{maples2017use}} and ensures anxiety induction is tested after a neutral baseline. Testing the reverse order was beyond our study’s scope.}.
We chose a seven-minute duration as a compromise between two bodies of work, {\cite{minns2019immersive}} {\cite{rowe1998effects}} which suggests intervals of around 1-5 minutes and {\cite{howarth1997occurrence}} which suggests intervals under 20 minutes (two 7-minute sessions and a 2-minute trial leading to 16 minutes). We settled on seven due to the lack of existing data-driven guidelines in the field.

The results of the self-reported anxiety through the Subjective Units of Distress scale (SUDs) showed significantly higher ratings ($p < 0.01$) in the anxious environment compared to the ratings in the relaxing environment using a Wilcoxon signed-rank test. 
In addition to self-reporting, our study revealed a statistically significant impact of virtual spiders on individuals' physiological measures (heart rate variability), providing strong support for our hypothesis. 
Further details of this experiment can be found in Appendix A.

One potential concern is that users may not have responded solely to the virtual spiders in the anxious environment, as the morgue setting itself may have contributed to their anxiety. The anxious environment was intentionally designed as an extreme case to elicit a clear physiological response. While the morgue setting may have influenced participants' reactions alongside the presence of spiders, our second study confirms that physiological measures can change specifically in response to virtual spiders. Therefore, we proceeded under the assumption that the virtual spiders would cause changes in physiological responses, which our second study further supports.

\section{Evaluating \framework~Framework Study}
\label{sec:Evaluating EDPCGRL Framework Study}

In this section, we cover an evaluation of our \framework~framework using an arachnophobia case study. 
In our \framework~framework, a Reinforcement Learning (RL) method adapts a virtual spider in real-time based on an individual's anxiety levels. We refer to our proposed method as the RL method in this section.
Our investigation aimed to address two main research questions: (1) whether our framework could elicit specific desired anxiety levels in participants, and (2) how effective our framework is compared to a more standard rules-based method.

To evaluate our framework, we conducted a human-subject study with the experimental design outlined in Section~\ref{sec: study2_precedure}.
This experimental design simulated a therapist setting two desired anxiety levels for an arachnophobic individual, increasing over time to achieve the desired exposure therapy effect. 
As such, we chose two desired anxiety levels, one low and one high, which allowed us to investigate whether our framework could induce more precise anxiety levels in participants compared to a rules-based method.
We note that a low-anxiety level followed by a high-anxiety level would be the opposite expected effect of a non-adaptive system, where a participant's anxiety response would normally decrease over time.
We give further evidence for this in Appendix B.7.

We compared our framework against a rules-based method, overviewed in Section~\ref{sec: rule-based}.
This study was approved by the University of Alberta, Ethics approval $Pro00113894-AME1$.
We ran a pilot study with three individuals which we used to set some parameters for the final study, which we identify below. 

\subsection{Rules-based Method}
\label{sec: rule-based}
Rules-based methods are currently the most common Artificial Intelligence approach for adaptive VRET applications~\cite{zahabi2020adaptive}. 
We implemented a rules-based method for our study, inspired by the work of~\citet{kritikos2021personalized}.
We selected the implementation by~\citet{kritikos2021personalized} since, to the best of our knowledge, it represents the only research on adaptive VRET for arachnophobia. 
We make a number of alterations to this approach to match our study design and overview these below.

Due to the nature of our study, we made the following adaptations and changes. Their agent is based on the concept of a ``correction factor'', representing the divergence between an individual's actual EDA and the desired EDA level. Following their approach, we defined a correction factor as the difference between the current anxiety level and the desired anxiety level divided by 10. This correction factor yields a value ranging from -1 to 1. 
They developed distinct formulas to determine spider attributes based on correction factors. We employed their formulas to determine the values for the virtual spider's attributes. Some attributes they defined for a virtual spider differ from ours. In Table~\ref{Table:rule-based}, we detail which of their attribute formulas corresponds to each of our attributes, for instance, their Size attribute was used for our Largeness attribute.
Notably, we mapped their Size attribute to both our Largeness and Color because they did not consider Color an attribute, prompting us to choose Size as the most suitable alternative given its ties to a spider's appearance.
To align with our RL method, we discretized these attributes as shown in Table~\ref{Table:rule-based}. 
We recognize the possibility of implementing both the rules-based method~{\cite{kritikos2021personalized}} and the RL method~{\cite{guzdial2019friend}} to make multiple changes concurrently.
However, due to RL safety considerations~{\cite{gu2022review}}, to avoid any potential risk of causing harm to participants, we opted to alter a single attribute at a time. In addition, this would allow us to better measure the impact of each change. Therefore, we adapted the rules-based method to randomly select one attribute from those originally chosen for modification.

\textbf{Example: } To better illustrate the differences between the rules-based method and our RL method, consider the following scenario where the target anxiety level is set to 7 during the high-anxiety section. The spider’s attributes are ordered as shown in Table {\ref{Table: spider_attr}}: [Locomotion, Amount of Movement, Closeness, Largeness, Hairiness, Color].

\textit{Rules-Based Method:}
The rules-based method begins with the spider [1,0,1,1,1,1], inducing an anxiety level of 4. It follows a set of predefined rules to modify an attribute but randomly selects the attribute from a pool of options based on those rules.
To increase the anxiety, the rules-based method selects the Amount of Movement attribute to increase, resulting in a spider with attributes [1,1,1,1,1,1], which induces an anxiety level of 5.
The method then increases the Locomotion attribute, leading to spider attributes [2,1,1,1,1,1]. The induced anxiety level remains at 5. This time, it chooses the Amount of Movement attribute and increases it, resulting in a spider with attributes [2,2,1,1,1,1], which induces an anxiety level of 7. Although this method reaches the target anxiety level of 7, it does not use prior experience but instead relies on pre-authored rules and random chance.

\textit{RL Method:}
The RL method starts with the spider [1,1,1,0,0,1], inducing an anxiety level of 4 and receiving a reward of 0.47.
Based on this feedback and previous experience, the RL method increases the Amount of Movement, creating a spider with attributes [1,2,1,0,0,1], which induces an anxiety level of 6 and yields a reward of 0.93.
The RL method then adjusts the Locomotion attribute, resulting in attributes [2,2,1,0,0,1], which induce the target anxiety level of 7, earning a reward of 1.

This example demonstrates that the RL method learns which attributes most effectively influence the user's anxiety, allowing it to customize the experience for the individual rather than relying on generalized rules applicable to all users. While this scenario is slightly modified for clarity, we derived it from similar real logs from our experiment, highlighting the key differences between the two methods.

\begin{figure}[tb]
\centering
\includegraphics[width=0.9\columnwidth]{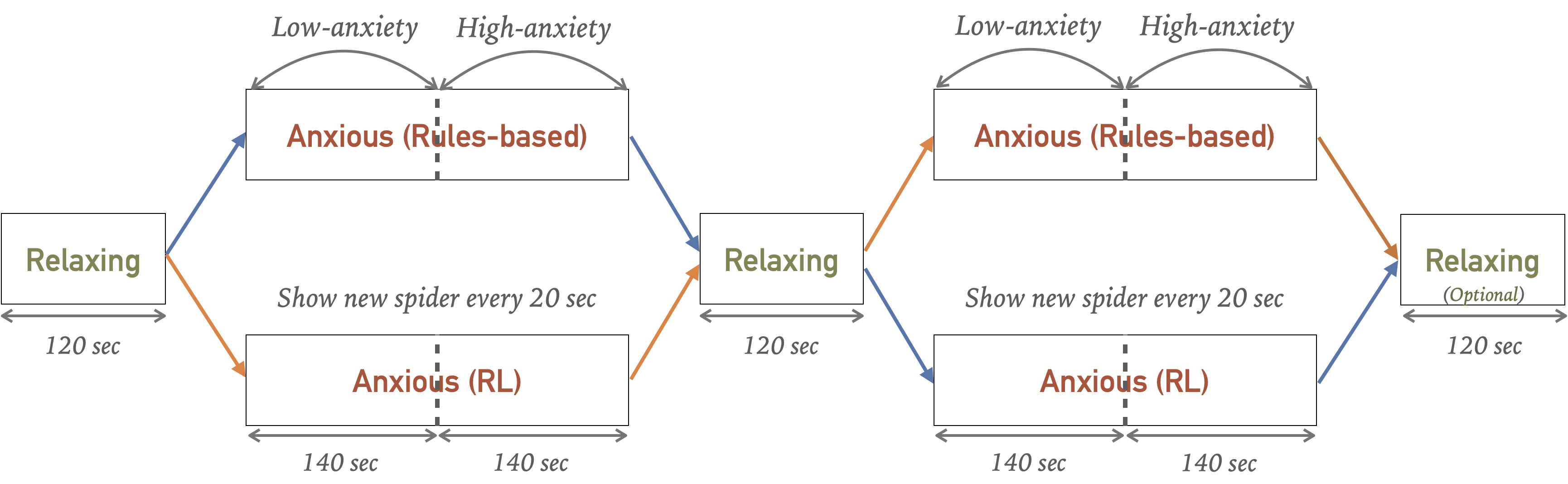}
\caption{An overview of two randomly counterbalanced assigned procedures in the human subject study. The order of environments within each procedure is distinguished by blue and orange colors.
Participants were first exposed to a relaxing environment for 120 seconds, followed by an anxious environment adapted by either the RL method or the rules-based method for 280 seconds. The desired anxiety level was low for the first 140 seconds and high for the next 140 seconds. Participants were then exposed to the relaxing environment again for 120 seconds. Finally, they were exposed to the anxious environment again using the alternative adaptive method, with the option to skip the final relaxing environment.}
\label{fig: procedure_study2}
\Description{The diagram displays three distinct components: Relaxing, Anxious (Rules-based), and Anxious (RL). The duration of each component is as follows: Relaxing lasts for 120 seconds, while both Anxious (Rules-based) and Anxious (RL) last for 280 seconds each. Each Anxious component is further subdivided into two segments: Low-stress and High-stress, with each segment lasting 140 seconds. Within the Anxious components, a new spider is introduced every 20 seconds. The figure outlines two sequences representing the two different study conditions, as indicated by separate arrows:
\begin{itemize}
    \item Relaxing > Anxious (Rules-based) > Relaxing > Anxious (RL) > Relaxing (optional)
    \item Relaxing > Anxious (RL) > Relaxing > Anxious (Rules-based) > Relaxing (optional)
\end{itemize}
}
\end{figure}
\subsection{Study Procedure}
\label{sec: study2_precedure}
We advertised the study through social media channels (specifically Telegram, Slack, and Discord) within our university given the requirements of an in-person study. 
To take part in the study, participants contacted the study coordinator over email, which was included in the social media posts.
Prior to participation, individuals were asked to complete a Fear of Spiders Questionnaire~\cite{szymanski1995fear} to ensure that they did not have arachnophobia. Those with arachnophobia were excluded from the study.
We made this choice to avoid causing undue harm,  given our absence of professional therapeutic qualifications.
In addition, we did not focus on treatment outcomes, only the ability of the system to adapt each individual to the desired anxiety levels.
We did not preclude participants who had taken part in the prior study, as the research questions were distinct and required different measurements.
We demonstrate the lack of significant impact on our results from participating in both studies via a statistical test in Appendix B.4. 

The study had to be in-person due to the need for a VR headset and a sensor to detect participants' physiological data. Participants received a \$10 CAD gift card for their participation.

Participants experienced one relaxing and two anxious environments, in one of two orders. The anxious environments were adapted to the user using either our RL method or the rules-based method. This study followed a within-subjects design.
The overview of the procedures is shown in Figure~\ref{fig: procedure_study2}. 
All participants ended with the same survey.

Upon arrival, participants were asked to read and sign an informed consent form. Then, we attached the physiological sensors to them. We chose electrodermal activity (EDA) sensors for this study using a Biopac device~\cite{biopac}. 
We selected EDA due to its well-established relationship with anxiety as discussed in Section~\ref{RelatedWork: Physiological}~\cite{senaratne2022critical}. 

We attached the EDA sensors to each participant's left palm which is known to have a high concentration of sweat glands~\cite{christopoulos2019body}. We decomposed the EDA signal into Skin Conductance Level (SCL) and Skin Conductance Response (SCR) using Smoothing Baseline Removal~\cite{braithwaite2013guide}. While there are other options beyond Smoothing Baseline Removal, we demonstrate in Appendix B.5 that they were largely equivalent. In this study, we specifically utilized the SCL data due to its positive correlations with anxiety characteristics~\cite{senaratne2022critical, cleworth2012influence, diemer2016fear, krupic2021anxiety}. The range of SCL measurements varies for each individual but typically falls between 1 and 20 microsiemens~\cite{braithwaite2013guide}. To address this variation, we define the minimum SCL value as the minimum observed SCL in the relaxing environment for each participant and assume a maximum value of 20. We then normalize this range to a scale of 0-10 for each participant, representing their level of anxiety for the purposes of our RL and rules-based methods.

Next, the participants were instructed to wear a VR head-mounted display and remain stationary throughout the experiment to mitigate any potential movement-related impacts on physiological recordings. 
Our choice for VR head-mounted display was the Oculus Rift S\footnote{https://www.oculus.com/rift/}. 
We selected the Oculus Rift S because it aligned with our study's needs including affordability, ease of installation and use, and the ability to connect to a computer for rendering high-fidelity graphics. Any VR head-mounted display meeting these criteria could have been used.

The participants were initially exposed to a relaxing environment for 120 seconds to stabilize their physiological measures. This 120-second duration was selected based on recommendations from studies indicating its adequacy for baseline EDA measurements~\cite{braithwaite2013guide, boucsein2012electrodermal} and is in line with the common baseline duration adopted in related research~\cite{king2023anxiety, pace2022eliciting, wilhelm2005mechanisms}. The relaxing environment was the same as our prior study (Figure~\ref{fig: relaxing}). 

\begin{figure*}[t]
  \centering
  \begin{subfigure}[b]{0.32\textwidth}
    \includegraphics[width=\textwidth]{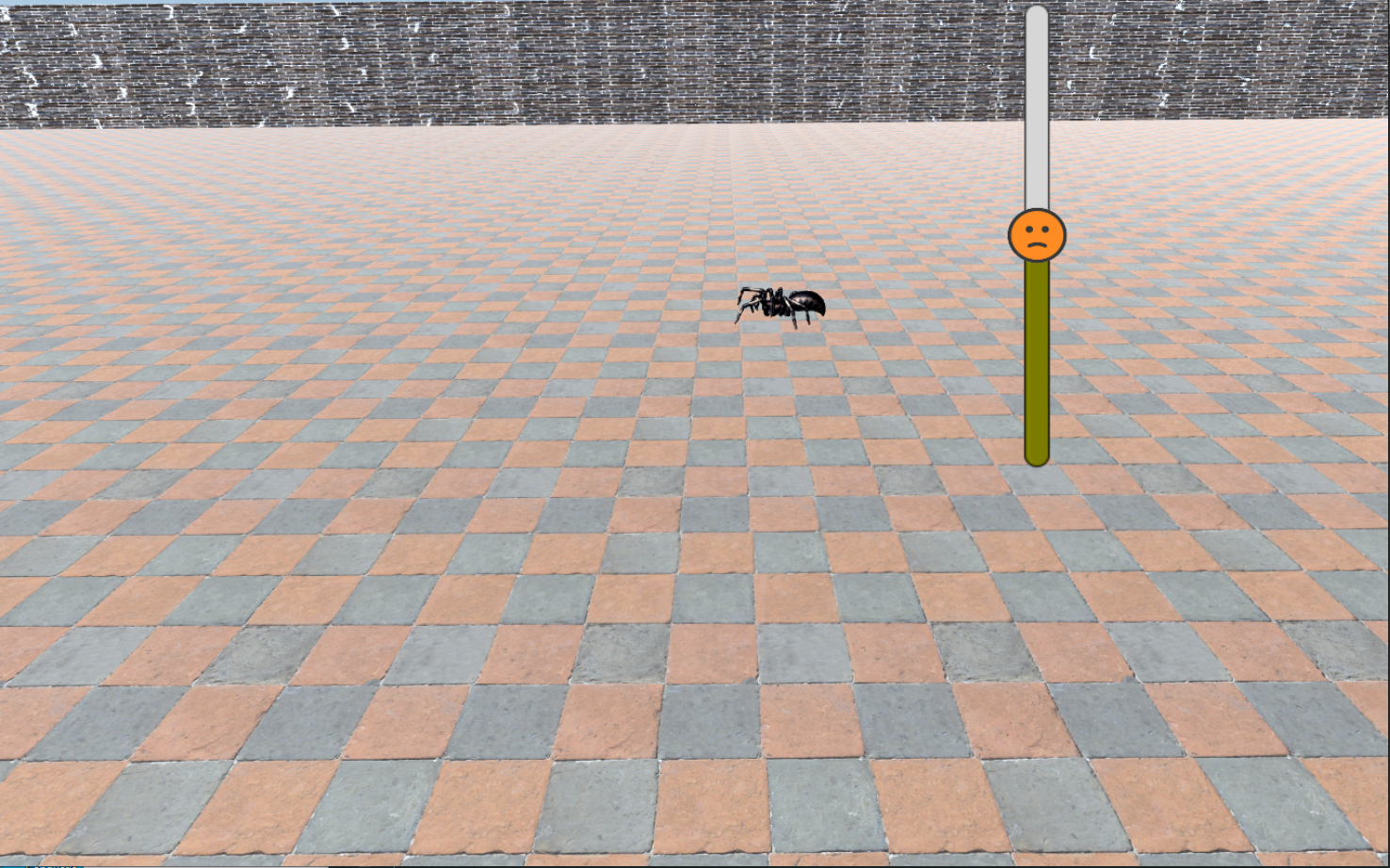}
     \caption{Minimum} 
     \label{fig: Minimum}
     \Description{When all the attributes of our virtual spider are set to their minimum possible values, it appears as a small, non-hairy, gray spider standing far from the participant. On the right side of the screen, there is a horizontal rating to collect SUDs.}
  \end{subfigure}
  \begin{subfigure}[b]{0.32\textwidth}
    \includegraphics[width=\textwidth]{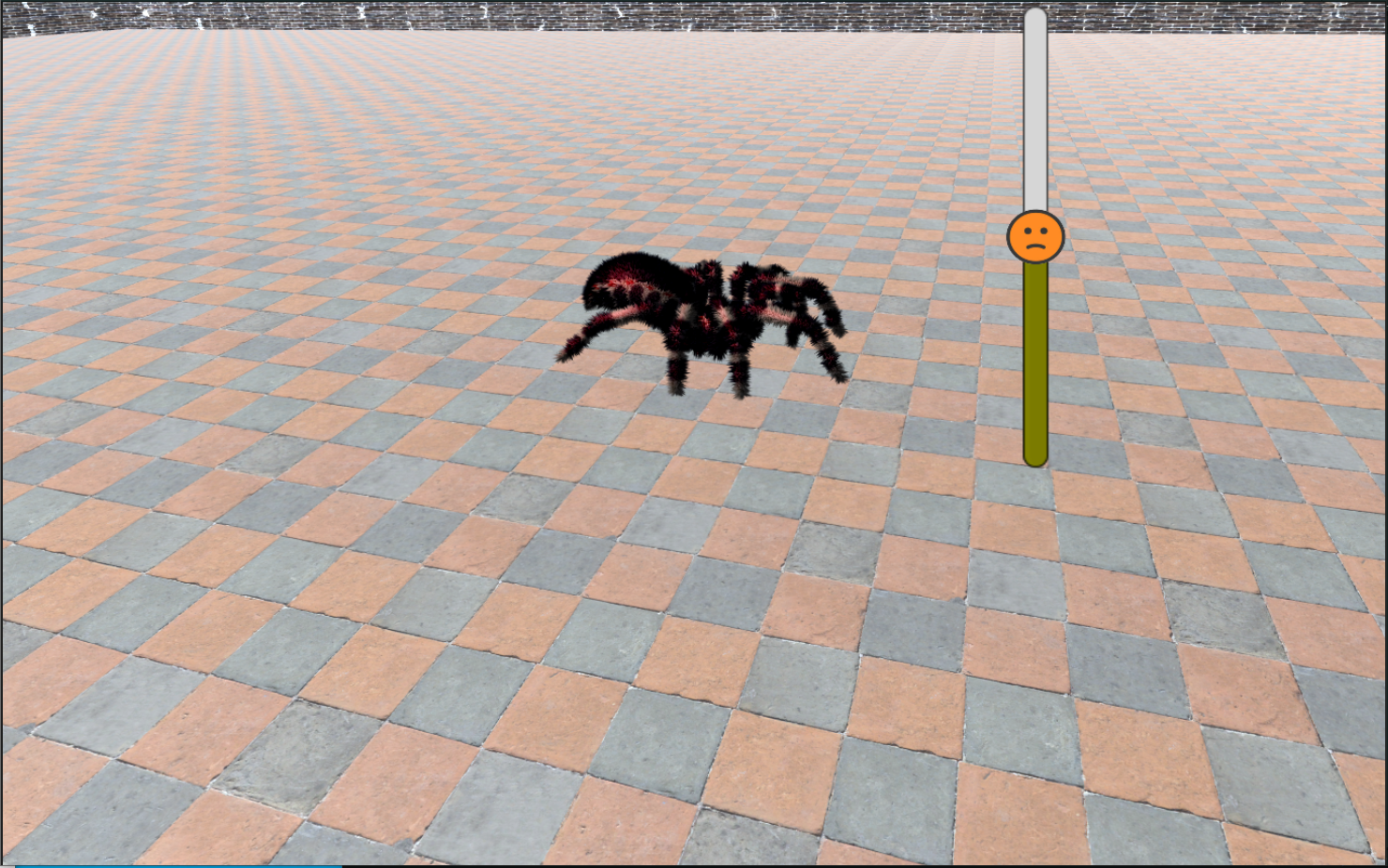}
    \caption{Average}
    \label{fig: Average}
    \Description{When all the attributes of our virtual spider are set to their average possible values, it appears as a medium-sized, hairy, red spider that walks a moderate distance from the participant. On the right side of the screen, there is a horizontal rating to collect SUDs.}
  \end{subfigure}
  \begin{subfigure}[b]{0.32\textwidth}
    \includegraphics[width=\textwidth]{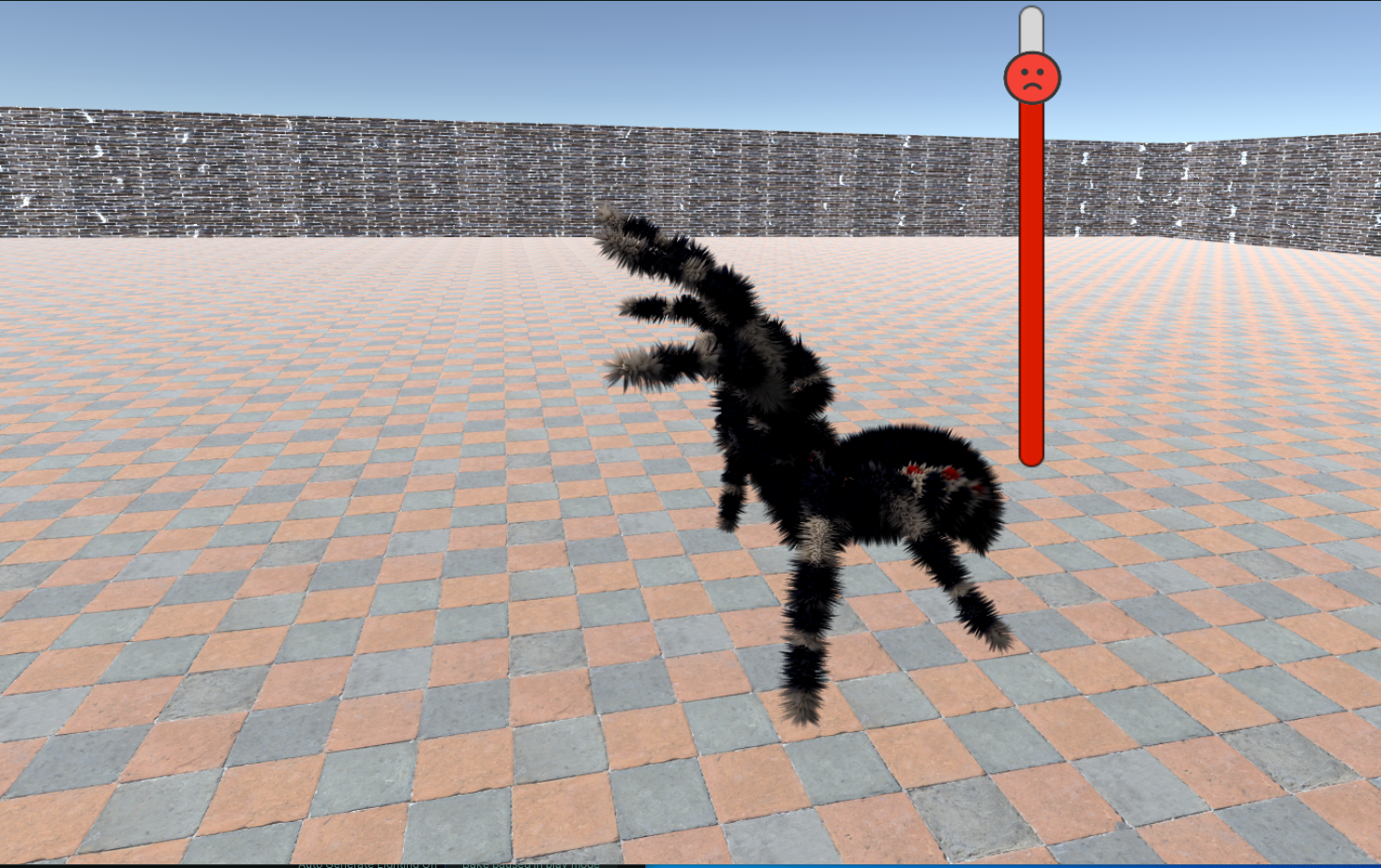}
    \caption{Maximum}
    \label{fig: Maximum}
    \Description{When all the attributes of our virtual spider are set to their maximum possible values, it appears as a large, hairy, black spider that moves quickly and jumps close to the participant. This spider in the picture is about to jump. On the right side of the screen, there is a horizontal rating to collect SUDs.}
  \end{subfigure}
  \caption{Generated spiders with three sets of attribute values.
  (a) When all attributes are set to their minimum values, the spider appears small, non-hairy, and gray, standing far from the participant. (b) With average attribute values, the spider is medium-sized, hairy, and red, walking at a moderate distance from the participant. (c) When all attributes are set to their maximum values, the spider becomes large, hairy, and black, moving quickly and jumping close to the participant.}
  \label{fig: VR_study2}
\end{figure*}

After exposure to the relaxing environment, participants were exposed to an anxious environment for 280 seconds, which included a spider modified based on one of our two adaptive methods. 
The adaptive method chosen for participants' initial exposure, whether our RL method or the rules-based method, was systematically counterbalanced. This ensured an equal distribution of participants experiencing either the RL method followed by the rules-based method or vice versa. This counterbalanced design was meant to mitigate potential biases that might arise from the order of method exposure.
This gave this study two conditions.
Regardless of the condition, the anxious environment began with a spider initialized with all attributes set to 0 (Figure~\ref{fig: Minimum}), as outlined in Table~\ref{Table: spider_attr}. The spider was adapted every 20 seconds based on the selected adaptive method. We selected a duration of 20 seconds based on our pilot study findings, striking a balance between preventing participant boredom and allowing sufficient time to observe changes in SCL.

\begin{table}
\small
\caption{The value of different spider's attribute using the correction factor for rules-based method based on~\citet{kritikos2021personalized}. We mapped the spider attributes from their study to ours and discretized these attributes, as they use a corresponding formulation that gives continuous numbers, while we consider spider attributes as discrete values.}
\label{Table:rule-based}
\begin{tabular}{llccc}
\toprule
       Our Attributes & Corresponding Formulation & 2 & 1 & 0 \\ 
\midrule
Locomotion &  Jumping Force & 
[-1 , -0.7) & [-0.7 ,  0.3] & (0.3 , 1]                                                    \\ 
Amount of Movement   &  Velocity &  
[-1 , -0.5) &  [-0.5 ,  0.4]  &  (0.4 , 1]    
                                \\ 
Closeness  &  Probability Moving Towards User &  
[-1 ,-0.7)  &   [-0.7 , -0.1]   & (-0.1 , 1] 
                                \\ 
Largeness &  Size  &  
[-1 , -0.8)  &   [-0.8 , -0.2]   & (-0.2 , 1] 
                                    \\

Hairiness  &     & 
   & [-1 , 0]  & (0 , 1]                            
                        \\ 
Color  &  Size & 
[-1 , -0.8)  &  [-0.8 , -0.2]   &  (-0.2 , 1]   
                        \\ 
\bottomrule
\end{tabular}
\end{table}

After the first anxious environment, the participants were placed in the relaxing environment for another 120 seconds followed by 280 seconds in the second anxious environment. In this second anxious environment, the spider adapted using the alternative method, either rules-based or RL, as seen in Figure~\ref{fig: procedure_study2}.  

During the first half of each of the anxious environments (0-140 seconds), labeled ``low-anxiety'', we set the desired anxiety level to 3 out of 10. In the second half (140-280 seconds), labeled ``high-anxiety'', the anxiety level was set to 7. Our aim was to determine if these adaptive methods could achieve anxiety levels in participants that matched the desired anxiety level.
We selected 3 and 7 as anchor points to ensure a sufficient spread of data for analysis, providing a meaningful range and distinction between the two levels while avoiding the extremes of 1 and 10.
Following the second anxious environment, the participants were exposed to the relaxing environment and had the option to either remain in it or skip it.

The participants were continuously asked to evaluate their Subjective Unit of Distress scale (SUDs) through a VR interface.
The rating scale was always visible, allowing participants to adjust their rating at any moment. 
We utilized this continuous, real-time rating system to provide us with a more comprehensive understanding of participants' experiences, similar to~\citet{srivastava2019continuous}.
We recorded these ratings every second. 

\subsection{Post-study Questionnaire}
After the VR experience, the participants were asked to fill out a questionnaire.
The questionnaire, presented via a Google Form, consisted of 19 questions.  

Questions 1-10 asked participants to reflect back on their VR experience, adapting questions from multiple existing questionnaires. 
The list of questions and their purpose is shown in Table~{\ref{Table: Questionnarie2_Q1_10}}.
Questions 11-19 included demographic questions and questions meant to identify confounding factors in our results (shown in Table~{\ref{Table: Questionnarie2_Q11_19}}). These questions were not adapted from prior questionnaires.
The questions were a mix of free response, ranking, 4-point Likert scale, 7-point Likert scale, checkboxes, and multiple choice questions. Our choice of 4-point or 7-point depended on what questionnaire we adapted the question from. We report the median value for Likert scale questions since the values were ordinal~\cite{sullivan2013analyzing}.
We did not conduct any analysis comparing the 4-point and 7-point Likert scale values; therefore, we feel confident in using medians to compare responses within each scale.

\begin{table}[]
\small
\caption{This table presents questions 1-10 from our post-study questionnaire, detailing each question, its purpose, and type. These questions prompt participants to reflect on their VR experience.}
\label{Table: Questionnarie2_Q1_10}
\begin{tabular}{p{0.4cm}p{6.3cm}p{4cm}p{1.8cm}}
\toprule
Num & Question                       & Purpose      & Answer Type    \\
\midrule
Q1, Q3      & How involved were you in the first/second virtual spider environment?          & 
Adapted from Presence Questionnaire (PQ)~\cite{whelansocial} to measure the sense of presence in VR.      & 7-point Likert \\
Q2, Q4      & Indicate the extent you felt ... while exposed to the first/second environment (first/second virtual spider environment): Calm, Tense, Upset, Relaxed, Content, Worried, Frightened, Uncomfortable, Nervous, Pleasant, Enjoyment, Frustration, Boredom, Disgust       & Derived from the State-Trait Anxiety Inventory (STAI)~\cite{spielberger1970manual} questionnaire to measure anxiety levels.                            & 4-point Likert \\
Q5          & Which of the environments did you find more...?
Pleasant, Stressful, Exciting, Fascinating, Interesting                                                                                                                     & Derived from a study by \citet{ninaus2019pilot}, designed for direct self-reported comparisons.  & Multiple-choice       \\
Q6, Q8      & Did you notice a change or adjustment in the spider while you interacted with the first/second environment
(the first/second virtual spider environment)?                                      & \multirow{2}{4.5cm}{Based on a similar study by \citet{bian2019design} to measure participants' conscious perception of changes in the spider.}   & Yes/No         \\
Q7, Q9      & If you noticed the change, then did you like how the system changed or adjusted for you?                                                                                &    & 7-point Likert \\
Q10         & Have you experienced any of the following symptoms while using virtual reality? general discomfort, fatigue, headache, eyestrain, difficulty focusing, increased salivation, sweating, nausea, difficulty concentrating, fullness of head, blurred vision, dizziness with eyes open/closed, dizziness with eyes closed, vertigo, stomach awareness, burping. & Derived from Simulator Sickness Questionnaire (SSQ)~\cite{kennedy1993simulator} to assess motion sickness and discomfort symptoms in VR.                             & Checkboxes    \\
\bottomrule
\end{tabular}
\end{table}
\begin{table}[]
\small
\caption{This table covers questions 11-19 of our post-study questionnaire, detailing each question, its purpose, and type. 
They include demographic questions and queries intended to identify confounding factors in our results.}
\label{Table: Questionnarie2_Q11_19}
\begin{tabular}{p{0.4cm}p{4.2cm}p{6cm}p{1.7cm}}
\toprule
Num & 
Question & 
Purpose & 
Answer type \\
\midrule

Q11 & 
Age & 
\multirow{3}{6.4cm}{Previous findings show their effects on EDA~\cite{bari2020gender, bari2020simultaneous}, and impacting perceived presence in VR~\cite{lorenz2023age, felnhofer2012virtual, lachlan2011experiencing} and VR sickness symptoms~\cite{macarthur2021you}.
} 
& 
Multiple-choice \\

Q12 & 
Gender & 
& Multiple-choice
 \\

&
&
&
\\

Q13 & 
How often do you play video games? & 
Prior gaming experience influences VR presence~\cite{lachlan2011experiencing} and VR sickness symptoms~\cite{weech2020narrative}.                                        & Multiple-choice       \\
Q14     & How would you rank your experience with Virtual Reality?           & Prior VR exposure affects individuals' self-perception and enjoyment within VR contexts~\cite{sagnier2020effects}.                                                                             & Multiple-choice       \\
Q15, Q17 & Are you afraid of any insects or bugs/spiders?                     &    Provide insights into participants’ fears related to arachnophobia.        & 7-point Likert \\
Q16     & If yes, which features of the insects/bugs scare you more?         & To understand
specific features of insects/bugs that participants thought would elicit fear.                                                        & Open-ended     \\
Q18, Q19 & When was the last time that you used stimulants/depressants today? & These substances can influence physiological measures~\cite{fletcher2017randomized, ronen2008effects, perez1982comparison}.                                                                    & Multiple-choice    \\
\bottomrule
\end{tabular}
\end{table}

\subsection{Participants}
22 non-arachnophobic participants (11 Males, 11 Females) aged 18-35 participated in our experiment. 
Concerning their gaming habits: 2 participants reported playing video games daily, 8 weekly, 4 monthly, and 8 less frequently (less than once a month), indicating a diverse range of gaming experiences among the group. Additionally, 7 participants had no prior experience with virtual reality, while 15 participants had some previous exposure to VR, with 13 of them not using it regularly.
The participants were assigned in a counterbalanced manner to either initially experience the RL method or the rules-based method.


Questions 15 and 17 of the questionnaire asked about fear of bugs/insects and spiders. 
The median response for both was 3.5 on a scale of 7 (with interquartile ranges of 3 and 2, respectively), where 7 means extremely frightened. 
This suggested a moderate level of unease or fear among our participants towards bugs/insects and spiders. 
As a follow-up in Question 16, participants outlined specific features of insects/bugs that scare them more. We conducted a qualitative content analysis, grouping the responses into categories based on our interpretation. These categories included movement (13 out of 22), size (8 out of 22), hairiness (5 out of 22), jump (5 out of 22), and speed (2 out of 22). 
This analysis indicates that movement and size were primary stress inducers for participants which aligns with previous studies and the design of our adaptive methods~\cite{lindner2019so}.

\subsection{Results}

The study yielded three sets of results, each discussed separately. 
We detail the findings from the SUDs and EDA sensors in Section~\ref{sec:study2_suds} and Section~\ref{sec:study2_eda} respectively. We then cover the results from the first ten questions of the questionnaire in Section~\ref{sec:setudy2_questionnarie}. Additionally, we analyze the order of the adaptive environment in Section~\ref{sec:study2_order}.

\subsubsection{Order of Adaptive Methods}
\label{sec:study2_order}
We first investigated whether the order of adaptive methods for the two anxious environments, specifically the RL method applied before the rules-based method or the reverse, influenced the SUDs or SCL. This analysis ensures that our results reflect the true effect of the adaptive strategies rather than potential confounding effects of their sequence.
To analyze this, we used Multivariate Analysis of Variance (MANOVA), which allowed us to examine the combined effects of these methods on both SUDs and SCL simultaneously. This approach is more advantageous than separate univariate analyses, as it accounts for potential correlations between the dependent variables. Utilizing MANOVA, we treated the order of the adaptive methods as an independent factor. Then we assessed its impact on the dependent variables: the mean SUDs/SCL during the low-anxiety and high-anxiety segments for both RL and rules-based methods. The results showed that the order in which the methods were applied did not significantly affect either SUDs 
or SCL 
This suggests that the choice of order of the adaptive methods has no significant impact on our results.

\subsubsection{Subjective Results: SUDs}
\label{sec:study2_suds}
\begin{figure*}[t]
  \centering
  \begin{subfigure}[b]{0.46\textwidth}\label{fig: SUDs1}
    \includegraphics[width=\textwidth]{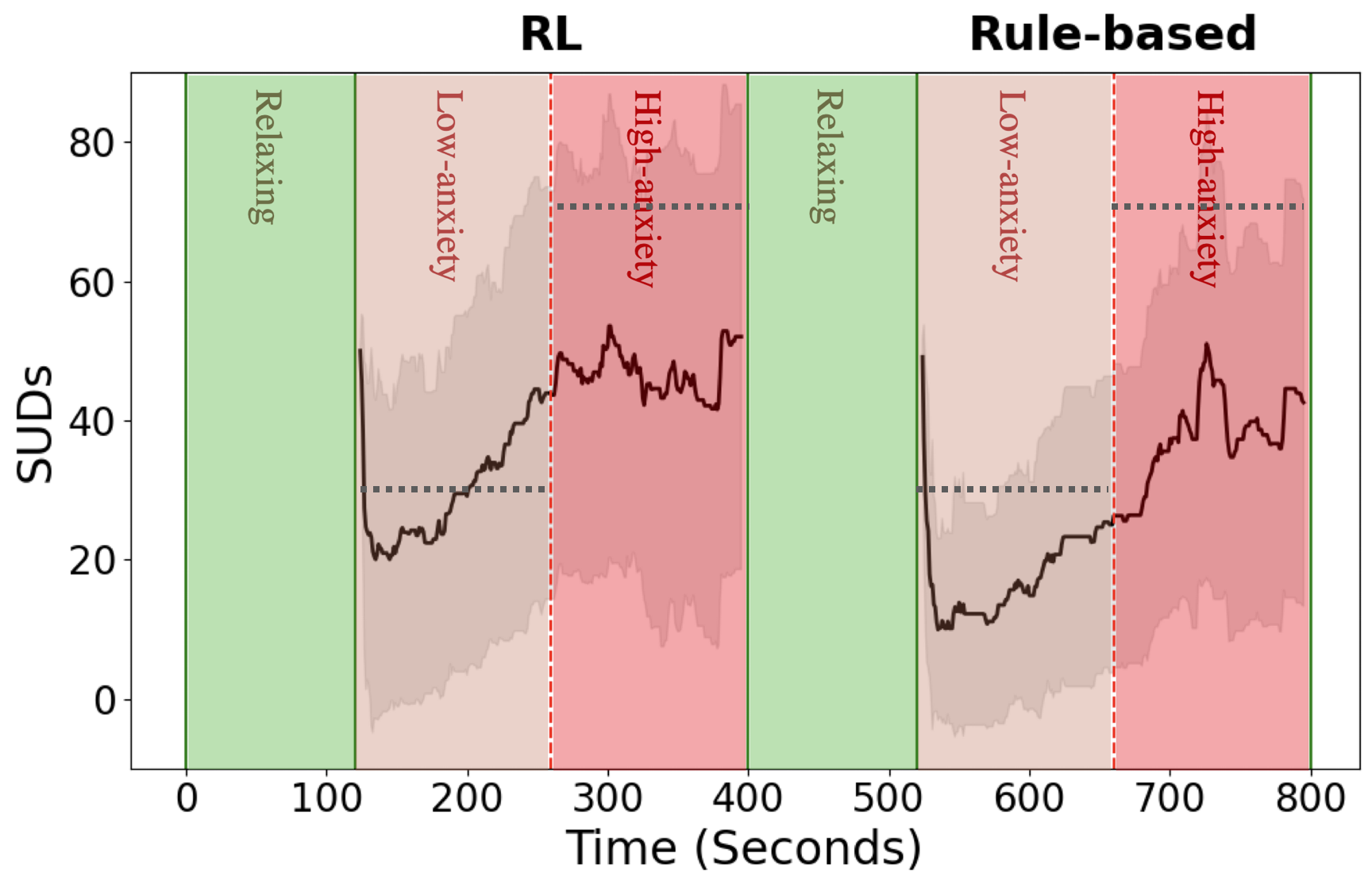} 
  \end{subfigure}
  \begin{subfigure}[b]{0.46\textwidth}\label{fig: SUDs2}
    \includegraphics[width=\textwidth]{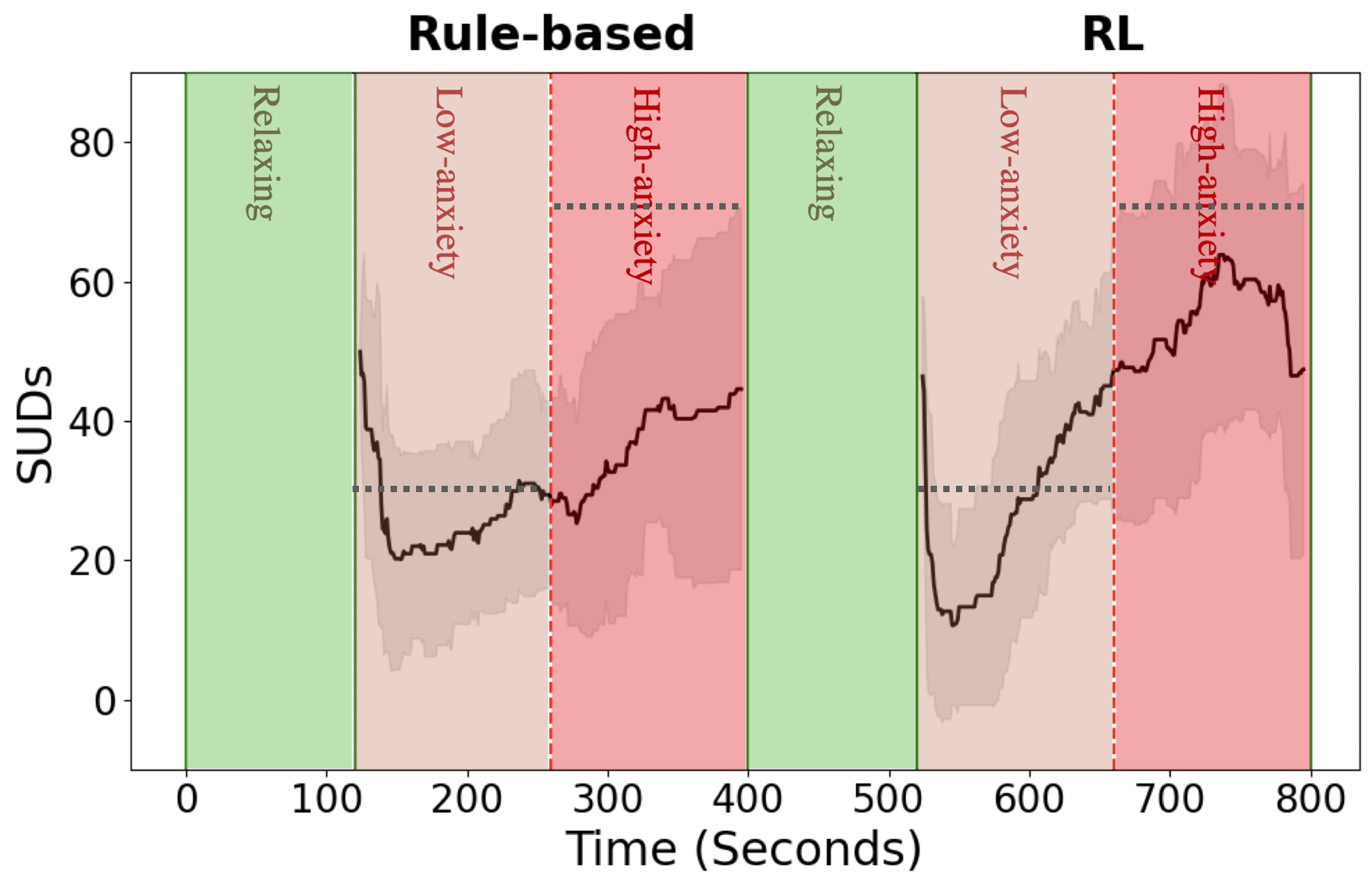}
  \end{subfigure}
  \caption{Average SUDs of participants during the VR experience. The left plot displays anxiety levels for those first exposed to an anxious environment adapted by the RL method, followed by the rules-based method. The right plot depicts the opposite order. The shaded region represents the standard deviation. We did not ask the participants to rate their SUDs during the relaxing environment. Note that the desired SUDs levels for Low-Anxiety and High-Anxiety are 30 and 70, respectively, as indicated by the dotted gray lines. Ideally, each adaptive method should produce a more pronounced increase in SUDs from low to high anxiety, with the resulting curves closely aligning with these target SUDs levels.}
  \label{fig: SUDs}
  \Description{This figure has two line graphs. Each line graph represents one of the two conditions (RL first or Rule-based first). Each line graph is segmented into six distinct sections: Relaxing (0-120 seconds), Low-anxiety (120-260 seconds), High-anxiety (260-400 seconds), Relaxing (400-520 seconds), Low-anxiety (520-660 seconds), and High-anxiety (660-800 seconds). 
  The x-axis represents Time (in seconds) and the y-axis shows the SUDs.
  As we did not collect SUDs during the Relaxing environment, these segments lack a corresponding line.
     
  The left graph shows the SUDs for the participants who were first exposed to the RL method. The SUDs values for each section are:
  Low-anxiety (M=30.36, SD=7.88, Min=20.0, Max=50.0), High-anxiety (M=46.87, SD=3.12, Min=41.55, Max=53.55), Low-anxiety (M=17.58, SD=6.06, Min=9.91, Max=49.09), and High-anxiety (M=37.38, SD=6.42, Min=25.55, Max=51.0). 

   The right graph shows the SUDs for participants who were first exposed to the rule-based method. The SUDs values for each section are:
   Low-anxiety (M=20.49, SD=5.97, Min=20.09, Max=49.91), High-anxiety (M=36.74, SD=5.84, Min=25.27, Max=44.55), Low-anxiety (M=27.42, SD=11.7, Min=10.64, Max=46.55), and High-anxiety (M=54.46, SD=5.61, Min=46.45, Max=63.82). }
\end{figure*}

We conducted a detailed analysis of the results of the participants' self-reported SUDs to determine (1) whether ratings during the higher anxiety level are significantly higher than those during the lower anxiety level, to confirm that the adaptive methods can effectively induce two distinct anxiety levels, and (2) the extent to which the reported SUDs levels aligned with the desired anxiety levels, to confirm that the adaptive methods can induce desired anxiety levels. This involved comparing the SUDs from both the low-anxiety and high-anxiety segments of the anxious environments. Additionally, we contrasted these findings between the environments adapted using either the RL method or the rules-based method.
Figure~\ref{fig: SUDs} showcases the SUDs data averaged over all participants.

\textbf{Individual Analysis:}
We first examined each participant to determine if their SUDs were higher when the desired anxiety level was higher. Given that the SUDs were not normally distributed, we opted for the Wilcoxon signed-rank test. The effect sizes were calculated using the formula $r = \frac{Z}{\sqrt{N}}$ proposed by \citet{fritz2012effect}. Notably, 19 participants ($\approx$ 86\% of participants) exhibited significantly elevated SUDs ($p < 0.01$, $r=0.64$) with the higher desired anxiety level and the RL method, compared to 16 ($\approx$ 72\% of participants) with the rules-based method ($p < 0.01$, $r=0.57$).
This demonstrates that both methods could influence SUDs values and show no significant difference (using a z-test for two proportions). However, the RL method exhibited marginally more consistency in influencing SUDs values.

\textbf{Aggregate Analysis:}
For each participant, we computed the mean SUDs for both the low-anxiety and high-anxiety segments within each anxious environment. 
The difference between these means over all participants was (M=21.77, SD=19.51) for the RL method and (M=15.02, SD=16.55) for the rules-based method. This mean difference is significantly higher for both the RL method ($t(21) = 5.23$, $p < 0.01$, $95\%~CI=[14.61, \infty]$) and the rules-based method ($t(21) = 4.26$, $p < 0.01$, $95\%~CI=[8.95, \infty]$) using a one-tailed paired t-test.
This larger difference suggests that the RL method may adapt to individuals and different anxiety levels more effectively compared to the rules-based method; however, it is not significantly better.

\textbf{Desired SUDs Analysis:}
In the prior analyses, we determined only if the SUDs were different between the low and high-anxiety segments. 
We now compare how precise the SUDs were compared to the desired anxiety levels.
Recall that our desired anxiety levels were 3 and 7 across the two halves of each anxious environment, on a 10-point scale.
Because of this, we assumed the desired SUDs to be 30 and 70 out of 100 possible values for the low and high-anxiety segments. We employed the Mean Squared Error (MSE) to compare actual SUDs with the desired ones, so a lower value is better. 

The RL method yielded average MSE values of (M=367.08, SD=464.74) and (M=891.49, SD=1309) for desired anxiety levels of 3 and 7, respectively. In contrast, the rules-based method had values of (M=273.30, SD=291.29) and (M=1538.12, SD=1484.52). This indicates that the rules-based method may have worked well for an anxiety level of 3, but it struggled with a high-anxiety level of 7. In comparison, our RL method demonstrated greater precision during high-anxiety segments.
The MSE values are significantly lower ($t(21) = 2.80$, $p < 0.01$, $95\%~CI=[249.89, \infty]$) in the high-anxiety segment for the RL method compared to the rules-based method using a one-tailed paired t-test, where lower means more precise. However, this difference is not observed in the low-anxiety segment. 

\textbf{Summary:} 
Both RL and rules-based methods could affect the anxiety of participants. However, the RL method demonstrated a stronger potential to effectively manipulate anxiety levels, particularly in achieving higher levels of anxiety.
\subsubsection{Objective Results: EDA Sensor}
\label{sec:study2_eda}
We conducted an in-depth analysis of the EDA data, mirroring the methodology applied to the SUDs. Therefore, the goal of this analysis is similar to the previous section: to determine (1) whether EDA data during the higher anxiety level are significantly higher than those during the lower anxiety level, in order to confirm that the adaptive methods can effectively induce two distinct anxiety levels, and (2) the extent to which the EDA data aligned with the desired anxiety levels, to confirm that the adaptive methods can induce desired anxiety levels.
While SUDs were self-reported values, the EDA data gave us an objective measure of participant anxiety responses.
For EDA data refinement, a low-pass filter was employed to minimize noise disturbances from electromagnetic fields or electrode contact inconsistencies~\cite{lee2020noise}. 

\begin{figure*}[t]
  \centering
  \begin{subfigure}{0.46\textwidth}\label{fig: EDA1}
    \includegraphics[width=\textwidth]{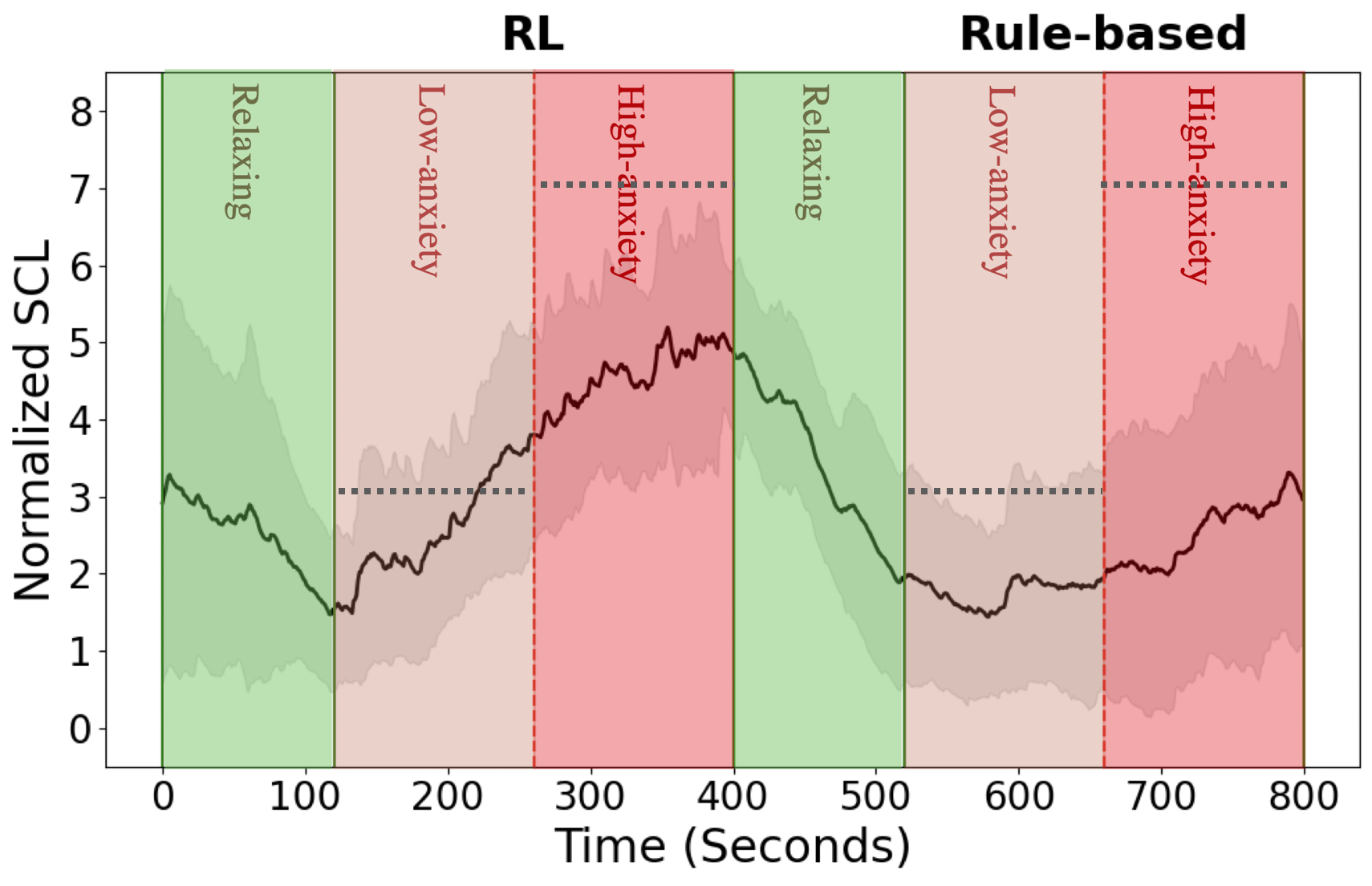} 
  \end{subfigure}
  \begin{subfigure}{0.46\textwidth}\label{fig: EDA2}
    \includegraphics[width=\textwidth]{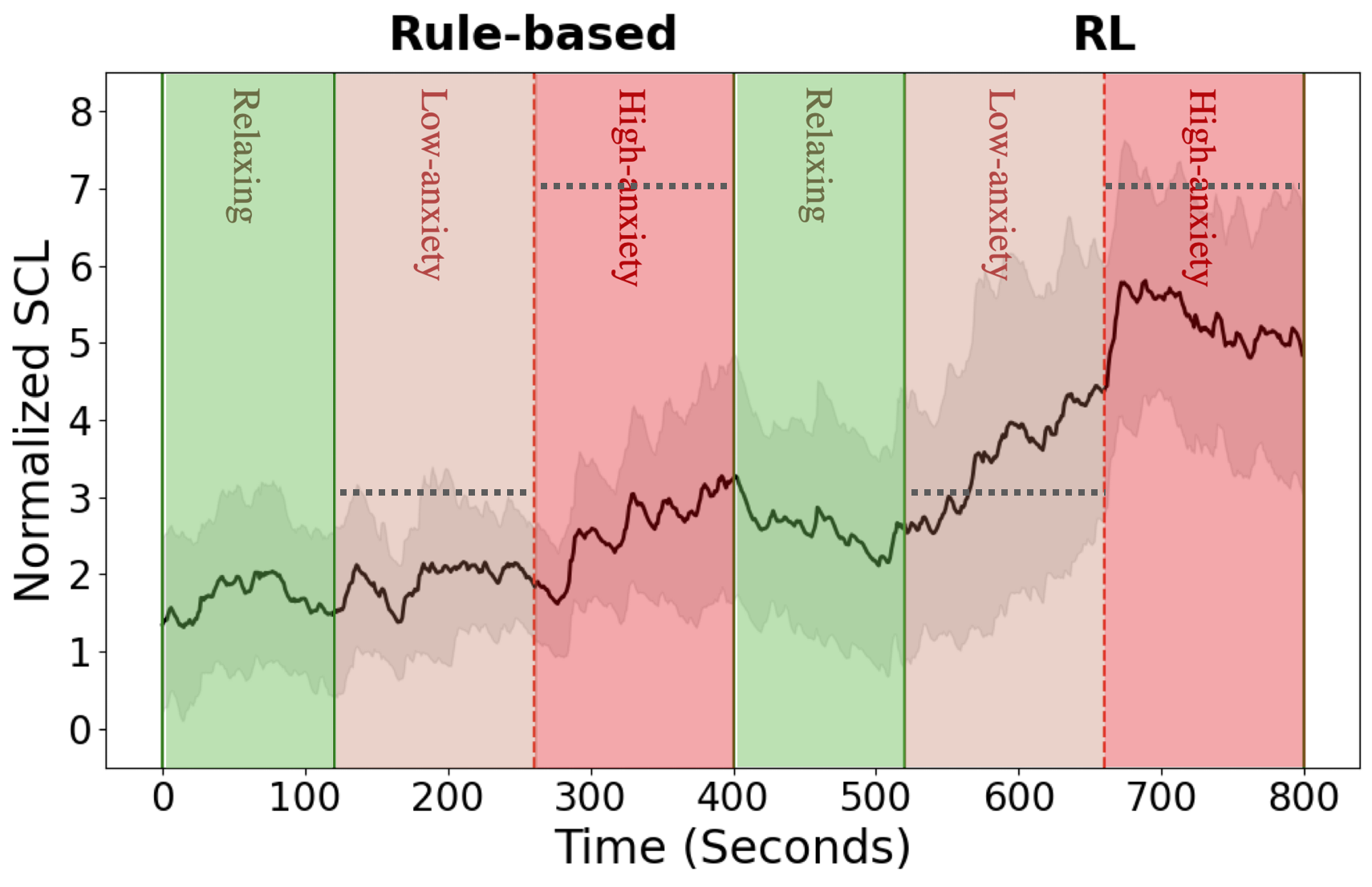}
  \end{subfigure}
  \caption{Average normalized SCL of participants during the VR experience. The left plot displays anxiety levels for those first exposed to an anxious environment adapted by the RL method, followed by the rules-based method. The right plot depicts the opposite order. The shaded region represents the standard deviation.
  Note that the desired SCL levels for Low-Anxiety and High-Anxiety are 3 and 7, respectively, as indicated by the dotted gray lines. Ideally, each adaptive method should produce a more pronounced increase in SCL from low to high anxiety, with the resulting curves closely aligning with these target SCL levels.
  On average, the rules-based method exhibited minimal changes between the low and high anxiety segments, suggesting it did not effectively adapt the spider to the individual participant's anxiety levels. In contrast, the RL method generally showed an increase in anxiety during the high anxiety segment, although it did not reach the desired anxiety levels consistently due to time constraints. This trend is further complicated by the right plot, where the RL method was experienced second, possibly affected by participant fatigue. Despite this, a slight upward trend toward the end of the period suggests that the RL approach could have achieved the desired anxiety level with more time.
Across both plots, the grey shaded area indicates that the RL method achieved higher anxiety levels for some participants in both conditions, whereas the rules-based method did not. 
  }
  \label{fig: EDA}
  \Description{This figure has two line graphs. Each line graph represents one of the two conditions (RL first or Rule-based first). Each line graph is segmented into six distinct sections: 
  Relaxing (0-120 seconds), Low-anxiety (120-260 seconds), High-anxiety (260-400 seconds), Relaxing (400-520 seconds), Low-anxiety (520-660 seconds), and High-anxiety (660-800 seconds). 
  The x-axis denotes Time (in seconds) and the y-axis displays the normalized SCL.
  
  The left graph line shows the SCL for the participants who were first exposed to the RL method. The SCL values for each section are:
  Relaxing (M=3.15, SD=0.6, Min=1.84, Max=4.11), Low-anxiety (M=3.2, SD=0.82, Min=1.87, Max=4.75), High-anxiety (M=5.72, SD=0.47, Min=4.71, Max=6.5), Relaxing (M=4.32, SD=1.17, Min=2.36, Max=6.1), Low-anxiety (M=2.22, SD=0.2, Min=1.8, Max=2.49), and High-anxiety (M=3.16, SD=0.51, Min=2.43, Max=4.14).

  The right graph line shows the SCL for the participants who were first exposed to the rule-based method. The SCL values for each section are:
  Relaxing (M=2.13, SD=0.27, Min=1.64, Max=2.55), Low-anxiety (M=2.4, SD=0.27, Min=1.72, Max=2.7), High-anxiety (M=3.21, SD=0.57, Min=2.03, Max=4.09), Relaxing (M=3.25, SD=0.31, Min=2.64, Max=4.09), Low-anxiety (M=4.4, SD=0.76, Min=3.17, Max=5.55), and High-anxiety (M=6.59, SD=0.39, Min=5.47, Max=7.25). }
\end{figure*}

We decomposed the EDA recordings to SCL and SCR as explained in Section~\ref{RelatedWork: Physiological}.
To ensure comparability across participants, SCL signals were normalized based on each individual's maximum value during the study. Specifically, we mapped each participant's minimum SCL value to 0 and their maximum SCL value (previously assumed to be 20) to 10. Given individual variations in baseline SCL, the actual observed maximum value across all participants was 8. This normalization process allowed for a standardized comparison of SCL changes between participants.
Figure~\ref{fig: EDA} showcases the SCL data averaged over all participants. The analysis is described in more detail below.

\textbf{SCL Individual Analysis:}
We first examined each participant to determine if their SCL recordings were more pronounced when the desired anxiety level was higher. Given the non-normal distribution of the SCL data, we used the Wilcoxon signed-rank test. All participants exhibited significantly increased SCL recordings ($p < 0.01$, $r=0.73$) during the higher desired anxiety level with the RL method. In contrast, only about 72\% of the participants (16 participants) with the rules-based method showed higher SCL during this period, though the difference was still significant ($p < 0.01$, $r=0.49$). This indicates that both methods could induce higher anxiety levels, but the RL method was significantly more consistent across participants (using z-test for two proportions, $p < 0.01$), compared to the rules-based method. 

\textbf{SCL Aggregate Analysis:}
For every participant, we computed the mean SCL during both the low-anxiety and high-anxiety segments within each anxious environment. Across all participants, the difference between these means was (M=2.28, SD=0.45) for the RL method and (M=0.94, SD=0.29) for the rules-based method. The difference between the low and high-anxiety values are significantly higher for both the RL method ($t(21) = 7.22$, $p < 0.01$, $95\%~CI=[1.74, \infty]$) and the rules-based method ($t(21) = 4.99$, $p < 0.01$, $95\%~CI=[0.62, \infty]$) according to a one-tailed paired t-test. 
However, the difference is again larger for the RL method.

\begin{figure*}[t]
  \centering
  \begin{subfigure}{0.3\textwidth} \label{fig: SCR_peaks}
    \includegraphics[width=\textwidth]{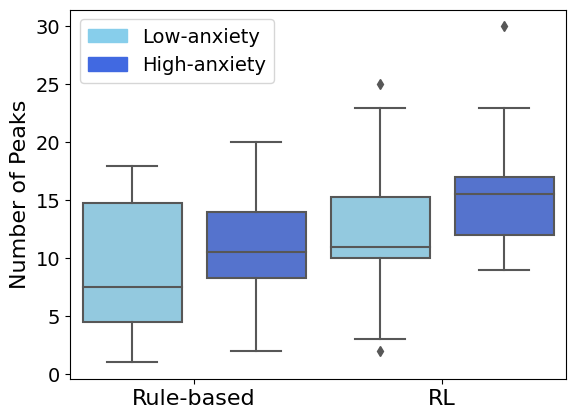} 
  \end{subfigure}
  \begin{subfigure}{0.3\textwidth} \label{fig: SCR_mean}
    \includegraphics[width=\textwidth]{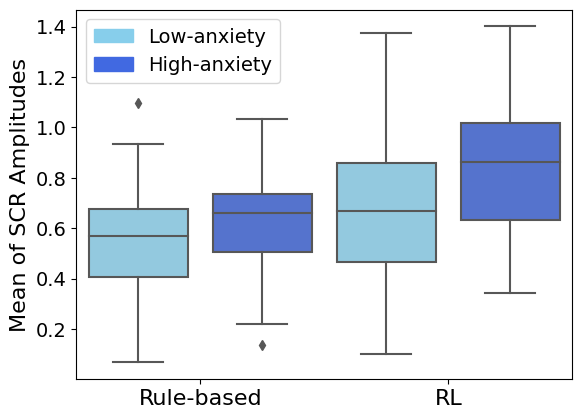}
  \end{subfigure}
  \begin{subfigure}{0.3\textwidth} \label{fig: SCR_sum}
    \includegraphics[width=\textwidth]{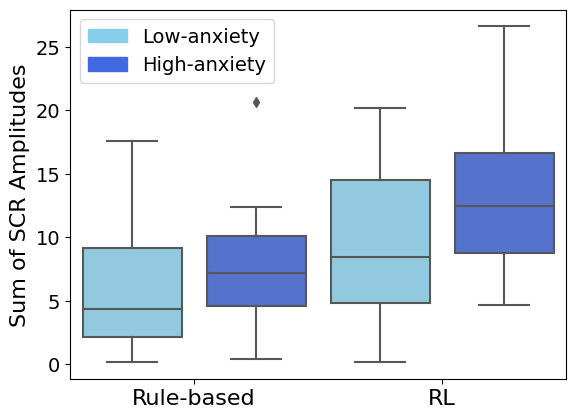}
  \end{subfigure}
  \caption{Comparison of three SCR features between the low-anxiety and high-anxiety segments when the adaptive methods were set to the rules-based method or RL method. The leftmost plot gives the number of peaks, the middle plot gives the mean of SCR amplitudes, and the rightmost gives the sum of SCR amplitudes. Across all three features, the RL method has a significant difference between the low and high anxiety values.}
  \label{fig: SCR}
  \Description{This figure comprises three boxplots, each representing four boxes: low-anxiety segments adapted by the rule-based method, high-anxiety segments adapted by the rule-based method, low-anxiety segments adapted by the RL method, and high-anxiety segments adapted by the RL method.

The left boxplot displays the number of peaks:
Low-anxiety and the rule-based method (M=9.27, SD=5.34, median=7.5),
High-anxiety and the rule-based method (M=10.86, SD=4.52, median=10.5),
Low-anxiety and the RL method (M=12.36, SD=5.85, median=11),
High-anxiety and the RL method (M=15.54, SD=4.81, median=15.5).

The middle boxplot shows the mean of the SCR amplitudes:
Low-anxiety and the rule-based method (M=0.55, SD=0.25, median=0.56),
High-anxiety and the rule-based method (M=0.61, SD=0.23, median=0.65),
Low-anxiety and the RL method (M=0.68, SD=0.31, median=0.67),
High-anxiety and the RL method (M=0.84, SD=0.28, median=0.86).

The right boxplot exhibits the sum of the SCR amplitudes:
Low-anxiety and the rule-based method (M=6.14, SD=4.98, median=4.37),
High-anxiety and the rule-based method (M=7.34, SD=4.65, median=7.21),
Low-anxiety and the RL method (M=9.14, SD=5.68, median=8.44),
High-anxiety and the RL method (M=13.12, SD=5.97, median=12.46).
  }
\end{figure*}

\textbf{SCR Aggregate Analysis:}
Prior studies have extracted the number of peaks and the mean or sum of SCR amplitudes to measure the intensity of anxiety~\cite{suso2019virtual, pick2018autonomic, yee2015insecure, albayrak2023fear, sevincc2018language}.
We extracted the same features from SCR to compare the intensity of anxiety between low and high-anxiety segments.
For each participant, we computed the number of SCR peaks, as well as the mean and sum of SCR amplitudes, during both the low-anxiety and high-anxiety segments within each anxious environment (shown in Figure~{\ref{fig: SCR}}). Given the normal distribution of these SCR features, we conducted one-tailed paired t-tests.
The number of peaks was notably higher for the RL method ($t(21) = 2.71$, $p < 0.01$, $95\%~CI=[1.38, \infty]$) during the high-anxiety segment compared to the low-anxiety segment. However, this is not true for the rules-based method. 
Similarly, the mean and sum of SCR amplitudes reflected the same trends. Both the mean and sum of SCR amplitudes were significantly higher ($t(20) = 3.68$, $p < 0.01$, $95\%~CI=[0.054, \infty]$ and $t(20) = 3.01$, $p < 0.01$, $95\%~CI=[1.31, \infty]$, respectively) in the high-anxiety segment for the RL method. However, the mean and sum of SCR amplitudes did not show significant increases 
in the rules-based method.
These findings indicated a significantly greater difference in SCR features (number of peaks, mean and sum of amplitudes) for the RL method, showcasing its efficacy in anxiety level modulation, outperforming the rules-based method.

\textbf{Desired SCL Analysis:}
We assumed the desired SCL to be 3 and 7 for the low and high-anxiety segments, aligning with our desired anxiety levels. 
We employed MSE to compare the actual SCL values with the desired ones, so again lower is better. The RL method yielded average MSE values of (M=3.31, SD=2.82) and (M=6.82, SD=4.92) for desired anxiety levels of 3 and 7, respectively. The rules-based method led to MSE values of (M=2.63, SD=2.26) and (M=22.34, SD=11.03). 
Both methods performed reasonably well at achieving the low-anxiety level, though the RL method was both more precise and consistent.
The MSE values did not show significant differences in the low-anxiety segment between the RL method and the rules-based method 
using a one-tailed paired t-test. However, the MSE values were significantly lower for the RL method ($t(21) = 7.51$, $p < 0.01$, $95\%~CI=[11.96, \infty]$) compared to the rules-based method in the high-anxiety segment.

\textbf{Summary:} 
Both the RL and rules-based methods can induce anxiety changes in participants. However, the RL method, as indicated by our EDA results, was markedly more capable of inducing a desired higher anxiety level. While the rules-based method also triggered anxiety changes, it did not achieve the same level of consistency or precision as the RL method.
This demonstrates that while the self-reported SUDs values might have indicated both approaches were comparable, the objective method suggests otherwise. 
This follows a typical pattern of difficulties individuals face when self-reporting their internal state~{\cite{inan2021method}}. 
However, the RL method also failed to consistently achieve the higher anxiety level for all participants, indicating space for future improvement.

\subsubsection{Questionnaire Results}
\label{sec:setudy2_questionnarie}
\begin{table}
\caption{The results of the questionnaire comparing two adaptive methods (rules-based vs RL) in adapting the virtual spider. The numbers indicate how many participants (out of 22) found one method to be more pleasant, stressful, exciting, fascinating, or interesting than the other, or both methods to be the same.}
\label{Table:compare_stressful}
\begin{tabular}{lccccc}
\toprule
More?      & Pleasant & Stressful & Exciting & Fascinating & Interesting \\ 
\midrule
RL         & 3        & 19        & 17       & 16          & 16          \\ 
Rules-based & 15       & 2         & 4        & 1           & 3           \\ 
Same       & 4        & 1         & 1        & 5           & 3           \\ 
\bottomrule
\end{tabular}
\end{table}

We now cover the non-demographic question responses.
Results from Question 5, contrasting the two anxious environments, are detailed in Table~\ref{Table:compare_stressful}. We counted the number of participants choosing each response and compared counts using a Binomial test. The findings suggested that a majority of the participants perceived the spider modified by the RL method as more stressful ($p < 0.01$). Participants reported finding the RL method more exciting, fascinating, and interesting, but not to a significant degree. Similarly, participants generally perceived the spider adapted via the rules-based method as more pleasant, but again not to a significant degree. 
This perception of ``pleasantness'' is relevant to the study because arachnophobia involves not only fear but also an exaggerated negative perception of spiders {\cite{kapustka2023analysis}}. Effective exposure therapy seeks to confront and reduce these distorted perceptions. If participants find the spider adapted via the rules-based method to be more pleasant, it may indicate that the method does not sufficiently challenge their fear responses, which is critical for achieving therapeutic progress.

\begin{figure*}[t]
  \centering
  \begin{subfigure}[b]{0.65\textwidth}
    \includegraphics[width=\textwidth]{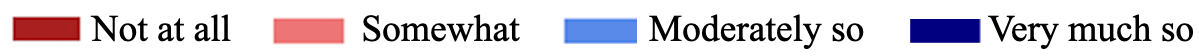}
  \end{subfigure}
  \begin{subfigure}[b]{0.48\textwidth}
    \includegraphics[width=\textwidth]{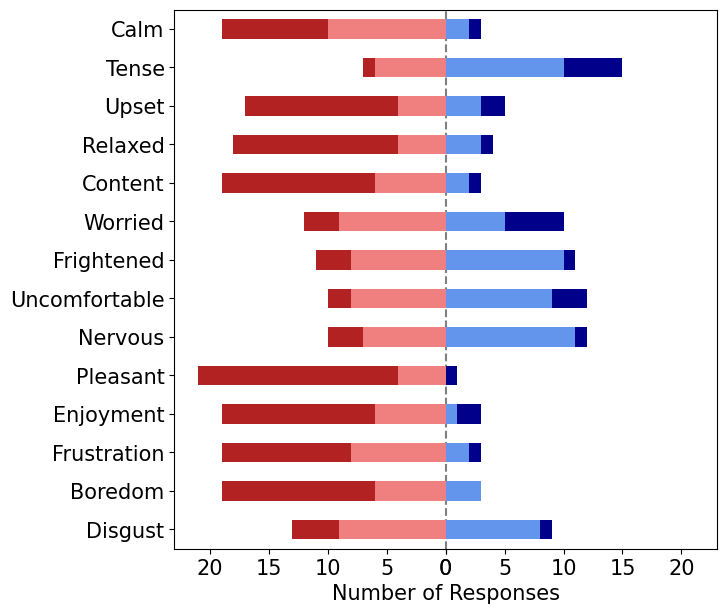}
    \caption{RL}
    \label{fig: STAI_RL}
    \Description{This figure displays a bar graph representing the extent to which participants felt each feature in the anxious environment that adapted using the RL method:
    \begin{itemize}
        \item Calm:
        	 Not at all (9 / 22); 
        	 Somewhat (10  / 22); 
        	 Moderately so (2 / 22); 
        	 Very much so (1 / 22)
        \item Tense:
        	 Not at all (1 / 22); 
        	 Somewhat (6  / 22); 
        	 Moderately so (10 / 22); 
        	 Very much so (5 / 22)
        \item Upset:
        	 Not at all (13 / 22); 
        	 Somewhat (4  / 22); 
        	 Moderately so (3 / 22); 
        	 Very much so (2 / 22)
        \item Relaxed:
        	 Not at all (14 / 22); 
        	 Somewhat (4  / 22); 
        	 Moderately so (3 / 22); 
        	 Very much so (1 / 22)
        \item Content:
        	 Not at all (13 / 22); 
        	 Somewhat (6  / 22); 
        	 Moderately so (2 / 22); 
        	 Very much so (1 / 22)
        \item Worried:
        	 Not at all (3 / 22); 
        	 Somewhat (9  / 22); 
        	 Moderately so (5 / 22); 
        	 Very much so (5 / 22)
        \item Frightened:
        	 Not at all (3 / 22); 
        	 Somewhat (8  / 22); 
        	 Moderately so (10 / 22); 
        	 Very much so (1 / 22)
        \item Uncomfortable:
        	 Not at all (2 / 22); 
        	 Somewhat (8  / 22); 
        	 Moderately so (9 / 22); 
        	 Very much so (3 / 22)
        \item Nervous:
        	 Not at all (3 / 22); 
        	 Somewhat (7  / 22); 
        	 Moderately so (11 / 22); 
        	 Very much so (1 / 22)
        \item Pleasant:
        	 Not at all (17 / 22); 
        	 Somewhat (4  / 22); 
        	 Moderately so (0 / 22); 
        	 Very much so (1 / 22)
        \item Enjoyment:
        	 Not at all (13 / 22); 
        	 Somewhat (6  / 22); 
        	 Moderately so (1 / 22); 
        	 Very much so (2 / 22)
        \item Frustration:
        	 Not at all (11 / 22); 
        	 Somewhat (8  / 22); 
        	 Moderately so (2 / 22); 
        	 Very much so (1 / 22)
        \item Boredom:
        	 Not at all (13 / 22); 
        	 Somewhat (6  / 22); 
        	 Moderately so (3 / 22); 
        	 Very much so (0 / 22)
        \item Disgust:
        	 Not at all (4 / 22); 
        	 Somewhat (9  / 22); 
        	 Moderately so (8 / 22); 
        	 Very much so (1 / 22)
    \end{itemize}
    }
  \end{subfigure}
  \begin{subfigure}[b]{0.48\textwidth}
    \includegraphics[width=\textwidth]{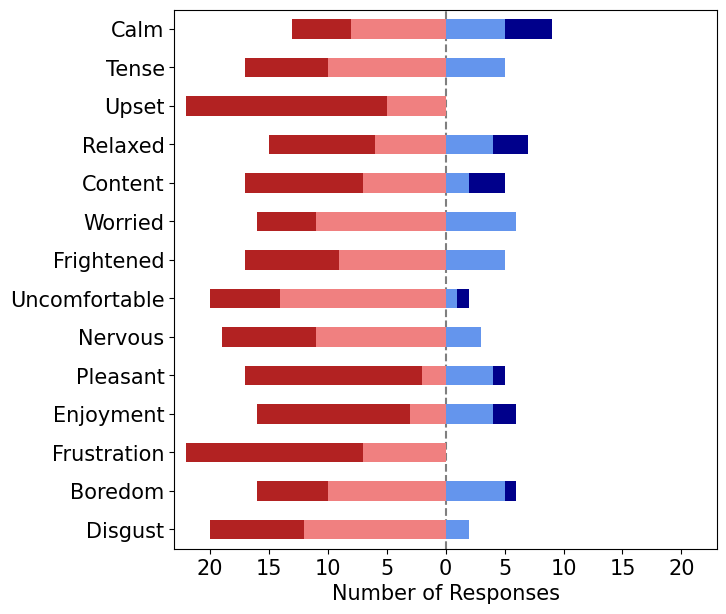}
    \caption{Rules-basaed}
    \label{fig: STAI_Rule}
  \end{subfigure}
  \caption{STAI results which show to what extent the participants felt each feature for the Evaluation EDPCGRL study. Based on these results, participants felt calmer, more relaxed, content, and pleased in the environment adapted using the rules-based method. On the other hand, they felt more tense, worried, frightened, nervous, and disgusted in the environment adapted using the RL method.}
  \label{fig: STAI_Study2}
      \Description{This figure displays a bar graph representing the extent to which participants felt each feature in the anxious environment that adapted using the rules-based method:
    \begin{itemize}
        \item Calm:
        	 Not at all (5 / 22); 
        	 Somewhat (8  / 22); 
        	 Moderately so (5 / 22); 
        	 Very much so (4 / 22) 
        \item Tense:
        	 Not at all (7 / 22); 
        	 Somewhat (10  / 22); 
        	 Moderately so (5 / 22); 
        	 Very much so (0 / 22)
        \item Upset:
        	 Not at all (17 / 22); 
        	 Somewhat (5  / 22); 
        	 Moderately so (0 / 22); 
        	 Very much so (0 / 22)
        \item Relaxed:
        	 Not at all (9 / 22); 
        	 Somewhat (6  / 22); 
        	 Moderately so (4 / 22); 
        	 Very much so (3 / 22)
        \item Content:
        	 Not at all (10 / 22); 
        	 Somewhat (7  / 22); 
        	 Moderately so (2 / 22); 
        	 Very much so (3 / 22) 
        \item Worried:
        	 Not at all (5 / 22); 
        	 Somewhat (11  / 22); 
        	 Moderately so (6 / 22); 
        	 Very much so (0 / 22)
        \item Frightened:
        	 Not at all (8 / 22); 
        	 Somewhat (9  / 22); 
        	 Moderately so (5 / 22); 
        	 Very much so (0 / 22)
        \item Uncomfortable:
        	 Not at all (6 / 22); 
        	 Somewhat (14  / 22); 
        	 Moderately so (1 / 22); 
        	 Very much so (1 / 22)
        \item Nervous:
        	 Not at all (8 / 22); 
        	 Somewhat (11  / 22); 
        	 Moderately so (3 / 22); 
        	 Very much so (0 / 22)
        \item Pleasant:
        	 Not at all (15 / 22); 
        	 Somewhat (2  / 22); 
        	 Moderately so (4 / 22); 
        	 Very much so (1 / 22)
        \item Enjoyment:
        	 Not at all (13 / 22); 
        	 Somewhat (3  / 22); 
        	 Moderately so (4 / 22); 
        	 Very much so (2 / 22)
        \item Frustration:
        	 Not at all (15 / 22); 
        	 Somewhat (7  / 22); 
        	 Moderately so (0 / 22); 
        	 Very much so (0 / 22)
        \item Boredom:
        	 Not at all (6 / 22); 
        	 Somewhat (10  / 22); 
        	 Moderately so (5 / 22); 
        	 Very much so (1 / 22)
        \item Disgust:
        	 Not at all (8 / 22); 
        	 Somewhat (12  / 22); 
        	 Moderately so (2 / 22); 
        	 Very much so (0 / 22)
    \end{itemize}
    }
\end{figure*}

Figure~\ref{fig: STAI_Study2} consolidates the responses to Questions 2 and 4, where participants rated the intensity of their experiences with each feature in the respective anxious environments. The observations reveal that participants felt the rules-based method provided a more calming, relaxing, and boring environment. In contrast, the RL method induced more tension, fear, discomfort, and disgust in the participants. Employing the STAI-6 scale, we quantified participants' anxiety levels on a spectrum from 20 (entirely relaxed) to 80 (highly anxious). 
Participants in the anxious environment with the RL method exhibited heightened anxiety levels (M=57.26, SD=11.83) compared to those in the rules-based method (M=46.2, SD=12.06). This difference in anxiety levels was statistically higher ($t(21) = 4.25$, $p < 0.01$, $95\%~CI=[6.58, \infty]$) for the RL method as determined by one-tailed paired t-tests.

\subsection{Results Discussion}
The main goal of this study was to assess the feasibility of our \framework~framework in inducing anxiety levels in participants that matched our desired anxiety level. We sought to compare its performance against a rules-based method adapted from \citet{kritikos2021personalized}. Our findings revealed that, while both methods influenced anxiety, our framework outperformed the rules-based method in terms of quantitative (EDA) and questionnaire results.

It is important to note that we tested the performance of the adaptive methods in a single session. 
Although we cannot make a definitive claim at this stage, we anticipate that the RL method may exhibit better performance over the long term. While part of this is just that this would give the RL method additional training data, it also has the opportunity to learn an individual's unique anxiety responses over time. This expectation arises from our results, which indicate that the RL method adapts more effectively to participants and seems to be converging to the desired values.  Further, given the results in Section~{\ref{sec: Spider Personalization Experiment}}, we conclude that our participants indeed experienced anxiety toward different types of spiders, emphasizing the necessity for individual adaptation.
For a comprehensive analysis of therapeutic outcomes, further assessment of the long-term impact is necessary.

We make three observations regarding the performance of the RL and rules-based methods in achieving the desired low and high anxiety levels in participants. Firstly, Figure~\ref{fig: EDA} shows that the RL method was able to reach the desired low and high-anxiety levels for both conditions. For the desired anxiety level of 3, on average it reached anxiety levels varying between 2 to 4. For the desired anxiety level of 7, on average it changed from 4 to 6, reaching the desired anxiety level of 7 for some participants (indicated by shaded areas in both plots). However, the rules-based method struggled to reach the desired anxiety levels, particularly for the high-anxiety level.
Secondly, neither of the methods consistently converged to the desired anxiety levels. This limitation may be attributed to intentionally limiting the time in each segment to mitigate common VR symptoms, as discussed in~\cite{howarth1997occurrence}. 
We identify from the left side of Figure~\ref{fig: EDA} that the RL method seems like it might have converged to the desired anxiety level in the average case given more time. 
Further experiments are required to investigate whether these methods can converge to the desired anxiety levels with an increased amount of time.

\section{Spider Personalization Analysis}
\label{sec: Spider Personalization Experiment}
Our prior study demonstrated the feasibility of our \framework~framework for personalizing spiders to achieve a specific anxiety response from target individuals. 
One potential criticism of our framework would be that there might just be a single spider across all individuals that could achieve each desired anxiety level (i.e., one low-anxiety spider, one high-anxiety spider, etc.). 
Instead, we hypothesized that different individuals would be more or less frightened by different attributes of our virtual spiders, which is consistent with prior arachnophobia research~\cite{lindner2019so}. 
We evaluated which spiders in our second study induced the highest anxiety response for each participant and then categorized them. 
If our hypothesis was correct, we anticipated that we would identify many distinct groups of spiders, representing differences across individuals. 
We first identified the spider that triggered the maximum anxiety response in each individual. 
We then employed K-Means clustering to group these spiders in terms of their attributes, based on prior work~\cite{alvarez2022toward, drachen2012guns, drachen2014comparison}. 
We identified eight clusters via the Elbow Method. 
We give the full results in Appendix C. 
We then associate each participant with the cluster that includes the spider that elicits the maximum anxiety response in them.
The resulting cluster centers were discretized to align with our spider attributes, as shown in Table~\ref{Table:Clusters}.

\begin{table}
\small
\caption{The spider attributes for each center of clusters and the number of participants who belong to each cluster. The data reveals that participants exhibited anxiety responses to spiders across all attribute values, except for Locomotion (no value 0 for Locomotion). Additionally, the diverse distribution of participants across clusters suggests a wide range of reactions, demonstrating that not all participants experienced the maximum anxiety response to the same type of spider.}

\label{Table:Clusters}
\begin{tabular}{cccccccc}
\toprule
Cluster & Locomotion & Amount of & Closeness & Largeness & Hairiness & Color & Number of \\
       &         & Movement   &         &         &         &      &        Participants    \\
\midrule
cluster 0 & 1          & 1        & 1         & 1         & 1         & 1     & 2                  \\
cluster 1 & 2          & 2        & 1         & 0         & 0         & 1     & 4                  \\
cluster 2 & 2          & 2        & 2         & 1         & 1         & 1     & 3                  \\
cluster 3 & 2          & 0        & 0         & 2         & 0         & 2     & 2                  \\
cluster 4 & 2          & 1        & 0         & 2         & 1         & 0     & 4                  \\
cluster 5 & 1          & 0        & 2         & 1         & 0         & 2     & 1                  \\
cluster 6 & 2          & 0        & 2         & 2         & 1         & 0     & 4                  \\
cluster 7 & 2          & 1        & 0         & 1         & 0         & 0     & 2                 
\\
\bottomrule
\end{tabular}
\end{table}

The table shows that different participants exhibited anxiety responses to spiders with all possible attribute values, except for Locomotion. Additionally, the distribution of participants across clusters was diverse, indicating a wide range of reactions. 
This demonstrates that not all participants showed the maximum anxiety response to the same type of spider.

We conducted an additional analysis to measure the anxiety induced by the spiders in each cluster for each of our participants.
Essentially, we measured how a participant, for whom the most anxiety-inducing spider appeared in one cluster, reacted when they saw a spider from another cluster.
If our hypothesis is correct, we would expect to see that each spider varied in terms of the amount of anxiety response from each participant, based on their cluster. 
For each participant, we measured the maximum anxiety level provoked by each spider in the clusters and then computed the mean values for all participants within each cluster. 
The results are presented in Table~\ref{Table:scares_clusters}, where \quotes{N/A} represents that participants in that cluster never saw any of the spiders in the other cluster. 
The table demonstrates that participants primarily showed an anxiety response to spiders from their own cluster.
Although they might exhibit a mild anxiety response towards one or two other clusters, their response generally remained less intense. For example, participants in cluster 3 were most distressed by spiders from the same cluster and showed some anxiety response towards cluster 5 spiders, but they did not exhibit a noticeable anxiety response towards other clusters.
These insights underscore that individuals' perceptions of the most anxiety-inducing spider can vary, emphasizing the need to personalize spiders to each person and avoid generalizations.
\begin{table}
\caption{The average of maximum anxiety level induced by the presenting spiders from different clusters. For example, the first row of values, shows the mean maximum anxiety experienced by participants in cluster 0 when exposed to spiders from different clusters. The color coding indicates the severity of anxiety, ranging from very light yellow (lowest) to red (highest). ``N/A'' indicates that participants in that cluster never encountered spiders from the other cluster. It demonstrates that participants primarily exhibited anxiety responses to spiders from their own cluster. Although they might show mild anxiety towards one or two other clusters, their response generally remained less intense.}
\label{Table:scares_clusters}
\begin{tabular}{l*{9}{c}}
\toprule
          & cluster 0 & cluster 1 & cluster 2 & cluster 3 & cluster 4 & cluster 5 & cluster 6 & cluster 7 \\
\midrule
cluster 0 & \heatmapcolor{7.5} & \heatmapcolor{3.5} & \heatmapcolor{4} & \heatmapcolor{4.5} & \heatmapcolor{4} & \heatmapcolor{4} & \heatmapcolor{4.5} & \heatmapcolor{4} \\
cluster 1 & \heatmapcolor{3.5} & \heatmapcolor{8.25} & \heatmapcolor{5.66} & \heatmapcolor{3.5} & \heatmapcolor{6.5} & \heatmapcolor{3.33} & N/A & \heatmapcolor{3.75} \\
cluster 2 & \heatmapcolor{5} & \heatmapcolor{2.5} & \heatmapcolor{7.33} & \heatmapcolor{5} & \heatmapcolor{6} & \heatmapcolor{2.33} & \heatmapcolor{5} & \heatmapcolor{2.66} \\
cluster 3 & \heatmapcolor{2.5} & \heatmapcolor{5} & \heatmapcolor{5} & \heatmapcolor{8} & \heatmapcolor{3} & \heatmapcolor{7} & \heatmapcolor{4} & \heatmapcolor{3} \\
cluster 4 & \heatmapcolor{6} & \heatmapcolor{3} & \heatmapcolor{6} & \heatmapcolor{6} & \heatmapcolor{7.5} & \heatmapcolor{0} & \heatmapcolor{6.5} & \heatmapcolor{5.25} \\
cluster 5 & \heatmapcolor{2} & \heatmapcolor{6} & \heatmapcolor{3} & \heatmapcolor{7} & N/A & \heatmapcolor{9} & \heatmapcolor{5} & \heatmapcolor{7} \\
cluster 6 & \heatmapcolor{5.75} & \heatmapcolor{5.5} & \heatmapcolor{5.75} & \heatmapcolor{6} & \heatmapcolor{7.5} & \heatmapcolor{5.33} & \heatmapcolor{8.25} & \heatmapcolor{6.75} \\
cluster 7 & \heatmapcolor{5.5} & \heatmapcolor{3} & N/A & \heatmapcolor{6} & N/A & N/A & \heatmapcolor{6} & \heatmapcolor{7}      
\\
\bottomrule
\end{tabular}
\end{table}

\section{Limitations and Future Work}
\label{sec:Limitations and Future Work}

Our research findings represent an improvement over the standard approach for automating personalized exposure therapy. However, it is crucial to acknowledge and address the limitations and concerns associated with this research.

A limitation of our study in Section{~\ref{sec:Evaluating EDPCGRL Framework Study} was relying solely on SCL as an anxiety indicator. Although EDA features have been shown to have a positive association with anxiety, they can be affected by both negative and positive emotions, e.g., amusement and anxiety{~\cite{senaratne2022critical}.
However, it is worth noting that the primary objective of the study was not to achieve perfect anxiety prediction.
In addition,
understanding the range of SCL for each individual can be difficult. One method to determine the range for a specific participant is to identify the minimum SCL when they are relaxed, and the maximum value when they experience a startle response. However, in this study, we chose not to startle the participants to avoid causing them further distress. 

Future research could benefit from incorporating additional sensors, like EEG and gaze data, for a more accurate estimation of anxiety levels. These sensors can offer unique perspectives and insights, such as identifying which aspects of the scene or spider participants focus on during the experiment.
While using multiple physiological measures 
could potentially enhance the accuracy of the anxiety level estimation, it would require additional sensors for participants, which may be costly and inconvenient. 

Furthermore, using machine learning (ML) techniques 
could improve anxiety level estimation.
However, ML methods rely on labeled datasets for training, necessitating precise knowledge of the induced anxiety levels, which can be challenging to determine as self-reporting by participants may not always accurately reflect their actual emotional state{~\cite{mandryk2006continuous}}.
Nevertheless, given our consistent results across both sensor and self-report readings, we believe we have successfully estimated anxiety in our studies.

While our current method is based on a basic RL technique, we are interested in exploring more advanced RL techniques to assess whether their performance exhibits notable variations. Given the simplicity of our state space, we do not anticipate a significant difference in performance or user experience with a more complex RL technique. However, more complex environments may benefit from these techniques. 

One other limitation is asking participants to remain stationary since movement could affect their physiological measures. To mitigate this issue, participants were given clear instructions to stay still and were monitored throughout the experiment. While we acknowledge that subtle movements, such as head rotation, can occur, we believe they did not significantly impact the results of our study. However, we recognize that this is a potential limitation.

Since VR is not commonly experienced in daily life, participants may experience excitement while using VR, potentially influencing their physiological sensor readings. Additionally, environmental factors, including caffeine intake, sleep patterns, and mood, can impact an individual's physiological responses. However, any such effects should remain consistent throughout the experiment, and we made every effort to minimize their impact.

In this research, no conclusions can be drawn about the health outcomes of our framework specifically for arachnophobic individuals for two reasons: (1) this was not the goal of our study, and (2) we selected arachnophobia as a case study but tested our framework on non-arachnophobic individuals. However, it is typical in studies like this to assume likely therapeutic efficacy based on physiological measures}{~\cite{hirai2007exposure, cote2009cognitive}}, we thus expect our framework—which outperformed the rules-based method— to demonstrate equal or superior efficacy. 
While our framework proved capable of adapting to non-arachnophobic individuals, it is reasonable to assume that it will perform equally well, or potentially better, for arachnophobic individuals due to their stronger reactions to virtual spiders, making adaptation easier. However, it should be noted that arachnophobic individuals might exhibit different behaviors, such as avoiding looking at the virtual spider, which might make the adaptation more difficult. Now that we have some evidence of the effectiveness of our framework, we hope to explore future work applications with arachnophobic individuals in collaboration with practicing therapists. 

From an ethical standpoint, any automated personalization poses concerns about producing unintentional distress or even re-traumatizing participants. To avoid this potential outcome, we designed our system to make minimal changes and to automatically stop if a participant reaches maximum anxiety. Future VR exposure therapy applications should be designed to avoid these potential negative outcomes. 

Our current framework uses basic safety measures, such as restricting attribute adjustments to incremental changes and setting predefined boundaries for spider attributes. However, our approach to safety considerations is still in the early stages and requires further exploration. RL systems used in clinical settings must ensure patient safety rigorously during both training (exploration) and deployment. Current methods, such as neurosymbolic RL{~\cite{anderson2020neurosymbolic}} and guidelines for trustworthy AI{~\cite{shneiderman2020bridging}}, emphasize the importance of formal verification methods and regulatory oversight for safe exploration. Therefore, future research should incorporate more robust safety measures, including formal verification techniques, human-in-the-loop control strategies, and constrained optimization approaches, to systematically ensure patient safety and compliance with ethical standards.

Furthermore, its implementation within a controlled laboratory environment, while valuable for initial validation, does not indicate our approach is currently ready for broader clinical applications. Additionally, the relatively small sample size may not fully capture the diversity of responses necessary to validate the framework across different populations and anxiety disorders. To address these limitations, we intend to partner with therapists who employ arachnophobia exposure therapy. This would begin with a collaborative, human-centered design approach to ensure that our system meets their needs. Following this, we would seek clinical trials with patients suffering from arachnophobia to validate the effectiveness of our system in terms of patient outcomes. In these trials, we would seek a broad range of participants across various demographic groups to ensure generalizability. The last step would be to obtain certification through a local health services agency to support widespread adoption.

}


\section{Conclusion}
\label{sec:Conclusion}

Enhancing the efficacy of Virtual Reality Exposure Therapy (VRET) often hinges on adapting to the unique needs of each individual instead of a one-size-fits-all approach.
In this work, we developed a novel framework for adaptive VRET using an Experience-Driven Procedural Content Generation via Reinforcement Learning
(EDPCGRL) approach.
We evaluated our framework via an implementation focusing on individuals with arachnophobia. We employed reinforcement learning in our implementation to modify a virtual spider in real-time based on a participant's anxiety indicators.
In our initial human subject study, we showed that our virtual spiders could elicit measurable anxiety reactions in users, as validated by subjective feedback and physiological data.
In our primary study, we evaluated the performance of our framework and compared it with a rules-based method.
Our results demonstrated that our framework can adapt virtual spiders to align with the desired anxiety levels in participants. 
Moreover, a significant variance was observed in user responses to different spider attributes, emphasizing the importance of a personalized therapeutic approach.
Returning to the motivating use case, we hope this framework can aid in providing personalized automated therapy that is more engaging and precise, but also less time-consuming for practitioners.

\section*{Acknowledgements}

This work was funded by the Canada CIFAR AI Chairs Program. We acknowledge the support of the Alberta Machine Intelligence Institute (Amii). We acknowledge the support of the Natural Sciences and Engineering Research Council of Canada (NSERC).

\bibliographystyle{ACM-Reference-Format}
\bibliography{main}

\end{document}